\documentclass[a4paper,fleqn]{cas-dc}

\usepackage[numbers]{natbib}

\usepackage{algorithm}
\usepackage{algorithmic}
\usepackage{graphicx}
\usepackage{subcaption}

\newtheorem{definition}{Definition}
\def\tsc#1{\csdef{#1}{\textsc{\lowercase{#1}}\xspace}}
\tsc{WGM}
\tsc{QE}
\tsc{EP}
\tsc{PMS}
\tsc{BEC}
\tsc{DE}

\begin{document}
\let\WriteBookmarks\relax
\def\floatpagepagefraction{1}
\def\textpagefraction{.001}

\shorttitle{}
\shortauthors{}

\title [mode = title]{Out-of-Distribution Detection in Heterogeneous Graphs via Energy Propagation}                      



%
\author[1]{Tao Yin}[style = chinese, orcid=0009-0000-4242-8193]


\ead{yintt@tju.edu.cn}
\credit{Conceptualization, Methodology, Software, Writing – original draft, Visualization}
\affiliation[1]{organization={School of New Media and Communication, Tianjin University},
    city={Tianjin},
    country={China}}

\author[2]{Chen Zhao}[style=chinese]
\affiliation[2]{organization= {Department of Computer Science, Baylor University},
    state={Texas},
    country={USA}}
\ead{chen_zhao@baylor.edu}
\credit{Methodology,  Writing – review \& editing,  Supervision}
\author[3]{Xiaoyan Liu}[style=chinese]
\affiliation[3]{organization={School of Qiyue Media and Communication, Cangzhou Normal University},
    city={Hebei},
    country={China}}
\credit{Validation, Writing – review \& editing}
\author[1]{Minglai Shao}[style=chinese]
\ead{shaoml@tju.edu.cn}
\cormark[1]
\credit{Writing – review \& editing, Supervision}

\cortext[cor1]{Corresponding author}



\begin{abstract}
Graph neural networks (GNNs) are proven effective in extracting complex node and structural information from graph data. While current GNNs perform well in node classification tasks within in-distribution (ID) settings, real-world scenarios often present distribution shifts, leading to the presence of out-of-distribution (OOD) nodes. OOD detection in graphs is a crucial and challenging task. Most existing research focuses on homogeneous graphs, but real-world graphs are often heterogeneous, consisting of diverse node and edge types. This heterogeneity adds complexity and enriches the informational content. To the best of our knowledge, OOD detection in heterogeneous graphs remains an underexplored area. In this context, we propose a novel methodology for OOD detection in heterogeneous graphs (OODHG) that aims to achieve two main objectives: 1) detecting OOD nodes and 2) classifying all ID nodes based on the first task's results. Specifically, we learn representations for each node in the heterogeneous graph, calculate energy values to determine whether nodes are OOD, and then classify ID nodes. To leverage the structural information of heterogeneous graphs, we introduce a meta-path-based energy propagation mechanism and an energy constraint to enhance the distinction between ID and OOD nodes. Extensive experimental findings substantiate the simplicity and effectiveness of OODHG, demonstrating its superiority over baseline models in OOD detection tasks and its accuracy in ID node classification. 
\end{abstract}



\begin{keywords}
Heterogeneous Graph \sep Out-of-distribution Detection \sep Energy
\end{keywords}

\maketitle

\section{Introduction}
\label{sec:introduction}
The rapid progression of graph neural networks (GNNs) has profoundly impacted various domains, where graph data play a crucial role. GNNs can extract rich structural information from graphs. This enables them to effectively model complex relationships in graph data~\cite{GraphStructure}. This capability has driven their widespread adoption across a diverse range of domains, including social networks, knowledge graphs, the world wide web, and numerous others. Notably, GNNs offer significant advantages in critical tasks such as node classification~\cite{GCN}, link prediction~\cite{zhang2018link}, and graph classification~\cite{GIN}, providing innovative solutions to long-standing challenges in these fields. 

\begin{figure}[!tbp]
  \centering
  \includegraphics[width=\linewidth]{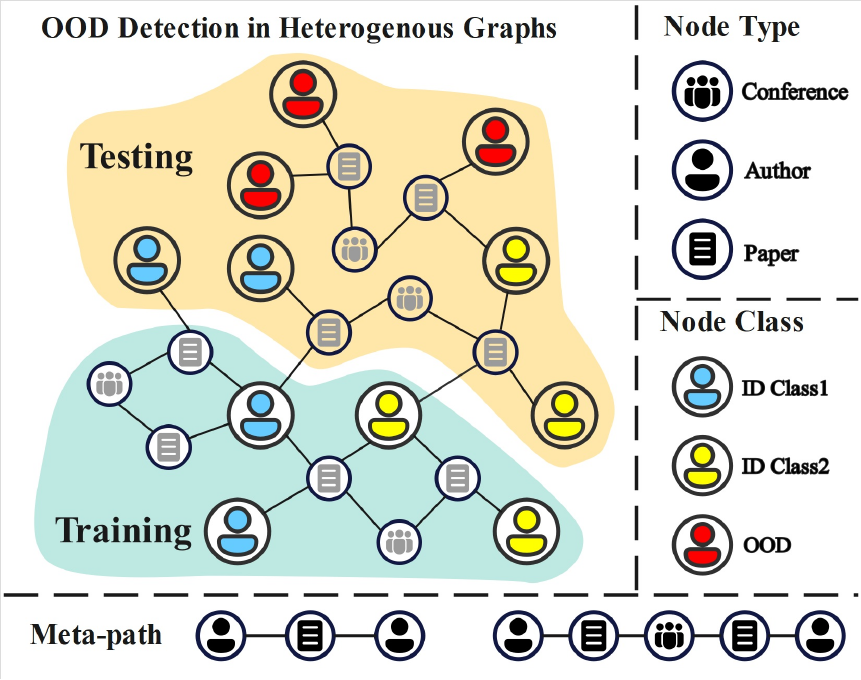}
  \caption{An illustration of OOD detection in heterogeneous graphs. The target nodes are authors, and during the training stage, we only utilize a subset of nodes from the ID classes (ID Class1 and ID Class2, as depicted in the figure). The task involves two main objectives: 1) detecting OOD nodes, and 2) classifying ID nodes.
 }
  \label{fig:1}
  \vspace{-3mm}
\end{figure} 

Despite the remarkable achievements of GNNs in numerous applications, a critical limitation remains: most existing research assumes that the training and test data originate from a common distribution~\cite{OpenSet}. Nevertheless, this assumption frequently does not hold in the real-world, where data frequently include out-of-distribution (OOD) samples~\cite{OODSurvey}. OOD samples are data points that deviate from the distribution of the training data, exhibiting discrepancies in features or labels. These samples can severely degrade model performance, resulting in unreliable predictions and impeding robust learning. Addressing this challenge is imperative for enhancing the generalization capability and reliability of GNN-based systems. While significant progress has been made in the areas of OOD generalization~\cite{survey3} and OOD detection~\cite{survey2}, these efforts have predominantly concentrated on image data, which are structured in regular, grid-like forms. In contrast, graph data are inherently irregular and non-Euclidean. They consist of interconnected nodes and edges that encode complex, often non-linear relationships. This fundamental structural difference necessitates the development of novel, specialised methods that are tailored to the unique properties of graphs. Existing OOD detection techniques, which are well-suited for grid-based image data, are inapplicable to graph-based tasks because of the irregularity and complexity of graph structures.

In recent years, several researchers have started investigating the detection of OOD samples in graphs, with a primary focus on homogeneous graphs. By leveraging message-passing mechanisms, researchers have successfully extracted structural information from homogeneous graphs. These approaches have led to promising results in OOD detection~\cite{EMP, GNNSafe}. However, real-world graphs are often heterogeneous~\cite{survey}. As graph data continues to be applied across diverse fields, such as social networks~\cite{social,social2}, recommendation systems~\cite{recommend,recommend2}, and citation networks~\cite{GraphSAGE,SimpleHGNN}, the heterogeneity of such data becomes increasingly prominent. In social networks, for instance, users may connect to various interest nodes, resulting in different types of edges. In recommendation systems, the relationships between users and items can take multiple forms, including purchases, views, and ratings. Similarly, in citation networks, nodes correspond to various elements such as authors, papers, and conferences, while edges denote relationships like authorship, citations, and publications. Heterogeneous graphs are formed by diverse node and edge types. Managing these varied types and their interactions poses challenges not found in homogeneous graphs~\cite{HGNN}. The additional complexity renders OOD detection methods developed for homogeneous graphs inapplicable to heterogeneous graph settings. In conclusion, conducting OOD detection in heterogeneous graphs is both more challenging and of greater practical significance. Figure \ref{fig:1} illustrates a real-world example of OOD detection in a heterogeneous graph. 

As we know, there is limited research that directly addresses the issue of OOD detection in heterogeneous graphs. Existing methodologies are predominantly devised for homogeneous graphs~\cite{OODGAT, GNNSafe}, overlooking the rich structural information provided by the diverse node and edge types in heterogeneous graphs. When applied on heterogeneous graphs, message-passing mechanisms designed for homogeneous graphs tend to focus on a single node type. This limitation makes it difficult to capture the complex relationships between different node and edge types. This leads to incomplete information propagation or the loss of critical node information. Such limitations significantly reduce the effectiveness of these methods in practical applications, as real-world graph data is inherently heterogeneous and rarely homogeneous. This has led an urgent need for novel approaches capable of accurately detecting OOD samples in heterogeneous graphs. These approaches should account for the unique interactions between various node and edge types, enabling more reliable and effective detection in real-world scenarios.

In this paper, we propose a novel method for OOD detection in heterogeneous graphs (OODHG). Unlike existing OOD detection methods that primarily focus on homogeneous graphs, our approach is specifically designed to capture the complex structural and semantic relationships in heterogeneous graphs. Our method first employs heterogeneous graph neural networks (HGNNs) to learn node representations while preserving multi-type node and edge dependencies. To quantify the likelihood of a node being OOD, we compute an energy score based on the logits produced by a multi-layer perceptron (MLP) layer. A key innovation of our approach lies in the introduction of a meta-path-based energy propagation mechanism, which refines energy scores by aggregating information from structurally and semantically relevant neighbors. This propagation step allows the model to better exploit the heterogeneous graph structure, leading to improved discrimination between OOD and ID nodes. Finally, OOD nodes are identified based on their refined energy scores, while ID nodes are classified using softmax probabilities. Compared to existing methods, our approach introduces a meta-path energy propagation mechanism, effectively addressing the limitations of homogeneous graph message-passing mechanisms when applied to heterogeneous graphs. This ensures that information transmission remains complete and prevents the loss of important node information.

To assess the efficacy of the proposed methodology, extensive experiments were conducted on three widely used heterogeneous graph datasets. The experimental results indicate that our method significantly outperforms existing baselines in OOD detection tasks in heterogeneous graphs. In particular, our approach achieves superior performance in both OOD detection and ID classification, demonstrating its robustness and practical relevance in real-world scenarios. This study not only provides a novel perspective on OOD detection in heterogeneous graphs but also lays the foundation for future research in this domain.

Our contributions can be summarized as follows:
\begin{itemize} 
\item To the best of our knowledge, this work represents one of the first systematic attempts to define and address the problem of OOD detection in heterogeneous graphs.
\item We propose a novel approach that integrates energy scores with the structural information of heterogeneous graphs. Through a meta-path-based energy propagation mechanism, our method efficiently detects OOD nodes. This approach leverages not only the node feature but also the rich structural properties of heterogeneous graphs. 
\item Extensive experiments on multiple heterogeneous graph datasets have been performed to demonstrate the effectiveness and wide applicability of the proposed methodology. \end{itemize}
    
\section{Related Work}
\label{sec:related}
\subsection{Heterogeneous Graph Neural Networks}

It has been demonstrated that GNNs can learn node representations by utilizing both the structural and feature-based information of the graph. Examples of such methods include GCN~\cite{GCN}, GAT~\cite{GAT}, and GraphSAGE~\cite{GraphSAGE}. Nevertheless, these techniques are created primarily for homogeneous graphs and exhibit limitations when extended to heterogeneous graphs. In homogeneous graphs, nodes and edges are of a single type. An example of a homogeneous graph is in social networks where nodes denote users and edges denote friendships. In contrast, heterogeneous graphs comprise a multitude of distinct node and edge types. To illustrate, in an academic network, nodes may represent various entities such as authors, papers, conferences, and research fields, while edges denote relationships such as authorship of papers or the association of papers with specific fields. This diversity within heterogeneous graphs provides richer information but also makes learning effective node representations more challenging.

HAN~\cite{HAN} introduces multiple meta-paths to capture distinct semantic information and employs a multi-level attention mechanism, combined with the GAT model, to capture the significance at both the node and semantic levels in heterogeneous graphs. This represents an early effort in addressing the challenges associated with heterogeneous graph analysis. GTN~\cite{GTN} automatically selects the most important meta-paths, transforming the input heterogeneous graph into meaningful meta-path graphs and learning node representations in an end-to-end way. MAGNN~\cite{MAGNN} encodes all information along a meta-path rather than focusing solely on the nodes at its endpoints, thereby addressing the issue of information loss present in HAN. Simple-HGN~\cite{SimpleHGNN} takes a path-free approach, whereby the features of disparate node types are mapped into a shared feature space. It employs a multi-layer GAT structure, combining node features with learnable edge-type embeddings to learn node representations. SeHGNN~\cite{SEHGNN} implements a monolayer structure that expands the receptive field through long meta-paths, while integrating features from multiple paths using a semantic fusion module.

\subsection{OOD Detection in Graphs}

In machine learning, OOD detection plays a crucial role in assuring model reliability and robustness in real-world applications. Extensive research has been conducted in fields such as computer vision and natural language processing, with approaches including MSP~\cite{MSP}, ODIN~\cite{ODIN}, DOC~\cite{DOC}, and Energy~\cite{Energy}. However, OOD detection presents unique challenges in the context of graph data, as none of these methods consider that graphs have non-Euclidean structures.

OOD detection in graphs can be separated into two tasks: node-level and graph-level. Graph-level OOD detection treats each graph as a sample, aiming to detect unseen graphs during the testing phase~\cite{graphood,graphood2}. Node-level OOD detection~\cite{he2025gdda,ma2025hypergraph,ding2025evidence,tian2024mldgg,wang2024feature,wang2024madod,shao2024supervised,zhao2023open,wang2022layer,zhao2019rank,zhao2021fairness}, which is the focus of this paper, treats each node as a sample and aims to detect nodes with unseen labels during the training phase. Existing methods for node-level OOD detection can be broadly categorized into two groups: uncertainty-based methods and structure-aware methods. 

Uncertainty-based methods: Wu et al.~\cite{OpemWGL} are pioneers in studying OOD detection on graphs. They introduce class uncertainty loss constraints to make the model more sensitive to unknown classes. Ultimately, they use a probability threshold to distinguish between ID and OOD nodes, providing a straightforward yet effective method for OOD detection on graph. Zhao et al.~\cite{GKDE} and Stadler et al.~\cite{GPN} expand on this by incorporating multiple types of uncertainty drawn from deep learning and evidence/belief theories. By integrating these uncertainties, their method provides a more robust framework for detecting OOD nodes.

Structure-aware methods: Song et al.~\cite{OODGAT} focus on the structural aspects of graph data, discovering that inter-class connections between ID and OOD nodes can hinder the accuracy of OOD detection. To address this, they propose an attention mechanism that selectively emphasizes relevant connections between nodes, improving the separation of ID and OOD nodes. Yang et al.~\cite{EMP} and Wu et al.~\cite{GNNSafe} take a different approach by using entropy and energy-based methods, respectively, to measure the probability that a node belongs to an out-of-distribution class. Chen et al.~\cite{chen2025decoupled} investigated the problem of OOD detection on heterophilic graphs and introduced an enhanced energy-based function to better handle the challenges arising from the heterophily of graph structures. Both works highlight the importance of leveraging graph structural information through propagation mechanisms, which help to improve the discrimination scores by spreading information across connected nodes, making OOD detection more robust.

\begin{figure*}[!tbp] 
  \centering
  \includegraphics[width=\textwidth]{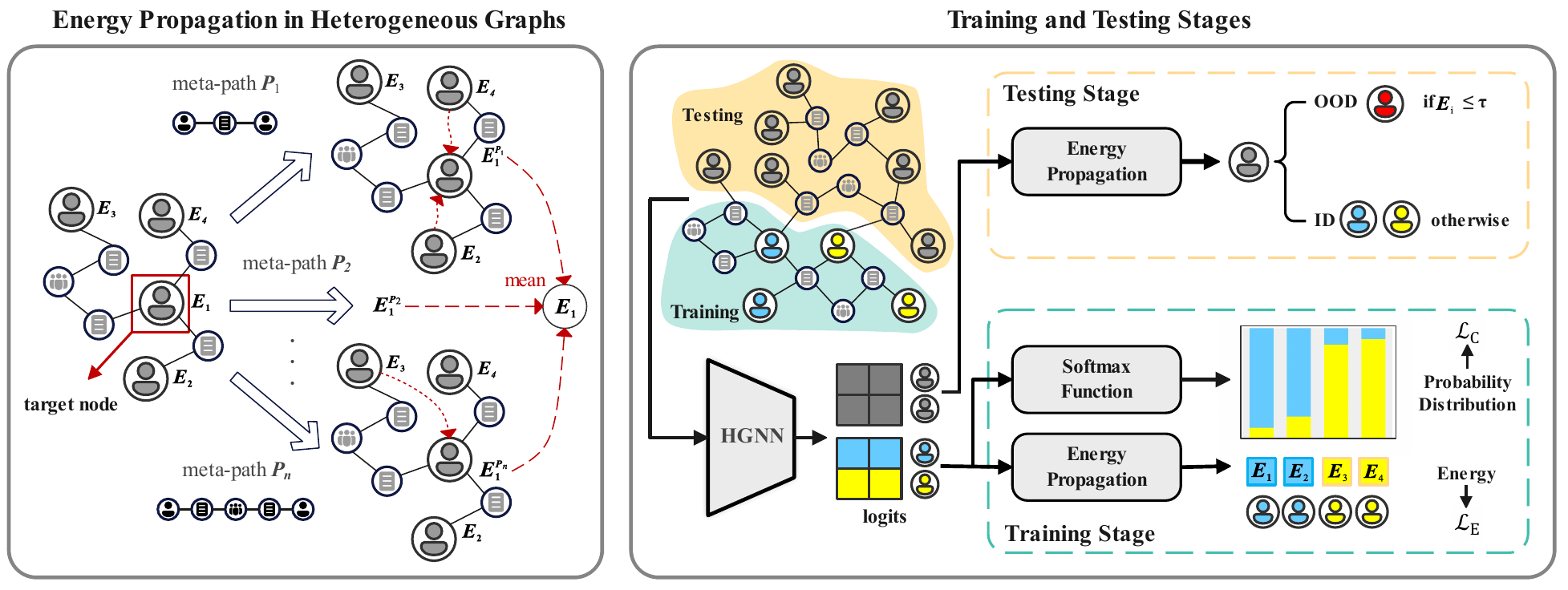}
  \caption{ The overall framework of OODHG. Given a heterogeneous graph \( \mathcal{G} \), HGNN outputs the logits \( H \) of the nodes and computes the energy scores for target type nodes. Energy propagation is performed under different meta-paths, and the average is taken as the final energy score. During the training phase, the final energy scores of target nodes in the training set are calculated for energy loss, while the classification probability distribution is used for classification loss. In the testing phase, the final energy scores for target nodes in the test set are calculated. If the negative energy score \(-E_i\) < \( \tau \), it is classified as an OOD node; otherwise, it is classified as an ID node.
 }
  \label{fig:2}
  \vspace{-3mm}
\end{figure*} 

\section{Preliminaries}
\label{sec:prelim}
In this section, we present the key definitions of heterogeneous graphs and the OOD detection problem within this context.

\begin{definition}[\textbf{Heterogeneous Graph}]
\textnormal{\( \mathcal{G} = \{ \mathcal{V}, \mathcal{E}, \phi, \psi \} \) represents a heterogeneous graph, where \( \mathcal{V} \) denotes the set of nodes and \( \mathcal{E} \) denotes the set of edges. Each node \( v_i \in \mathcal{V} \) is assigned a type\( \phi(v_i) \), while each edge \( e_{ij} \in \mathcal{E} \) is labeled with a type through \( \psi(e_{ij}) \). The types of nodes and edges are captured by the sets \( T_v = \{ \phi(v) : v \in \mathcal{V} \} \) and \( T_e = \{ \psi(e) : e \in \mathcal{E} \} \), respectively. When \( |T_v| = |T_e| = 1 \), the graph simplifies into a standard homogeneous graph. As shown in Figure~\ref{fig:1}, the heterogeneous graph contains different types of nodes: authors, papers, and conferences. There are edges connecting authors to papers and papers to conferences. The target type $t$ for node classification is authors, meaning that only the nodes representing authors are subject to classification.}
\end{definition}

\begin{definition}[\textbf{Meta-path}] 
\textnormal{A meta-path defines a composite relation of multiple edge types~\cite{metapath}, denoted as \( {P} = t_1 \rightarrow t_2 \rightarrow \ldots \rightarrow t_l \) (abbreviated as \( {P} = t_1 \ldots t_l \)), where \( t_i \) represents a node type. For example, as illustrated in Figure~\ref{fig:1}, two authors can be connected through multiple meta-paths, such as author \(\rightarrow\) paper \(\rightarrow\) author, and author \(\rightarrow\) paper \(\rightarrow\) conference \(\rightarrow\) paper \(\rightarrow\) author.}
\end{definition}
\begin{definition}[\textbf{OOD Detection in Heterogeneous Graphs}]
\textnormal{Given a heterogeneous graph \( \mathcal{G} = \{ \mathcal{V}, \mathcal{E}, \phi, \psi \} \) and a target type \( t \), the set of target nodes \( \mathcal{V}_t = \{ v \in \mathcal{V} \mid \phi(v) = t \} \) is partitioned into two disjoint subsets: \( \mathcal{V}_t = \mathcal{V}_t^{\text{train}} \cup \mathcal{V}_t^{\text{test}} \), where \( \mathcal{V}_t^{\text{train}} \) denotes the training node set and \( \mathcal{V}_t^{\text{test}} \) denotes the test node set. Each node \( v_i \in \mathcal{V}_t^{\text{train}} \) is associated with a label \( y_i \in \{ 1, \ldots, K \} \), where \( K \) represents the number of classes. The test set \( \mathcal{V}_t^{\text{test}} \) is further divided into two subsets: \( \mathcal{V}_t^{\text{test}} = \mathcal{V}_t^{\text{id}} \cup \mathcal{V}_t^{\text{ood}} \), where \( \mathcal{V}_t^{\text{id}} \) represents the set of ID nodes, which belong to the same distribution as those in \( \mathcal{V}_t^{\text{train}} \) and \( \mathcal{V}_t^{\text{ood}} \) represents the OOD nodes, which belong to the different distribution as those in \( \mathcal{V}_t^{\text{train}} \). OOD detection in heterogeneous graphs consists of two primary tasks: 1) identifying OOD nodes in \( \mathcal{V}_t^{\text{ood}} \), and 2) classifying the nodes in \( \mathcal{V}_t^{\text{id}} \). As depicted in Figure~\ref{fig:1}, the training set includes only InD nodes, while the goal is to detect OOD nodes in the test set and classify the ID nodes.}
\end{definition}

\section{Methodology}
In this section, we propose OODHG. An overview of OODHG is shown in Figure~\ref{fig:2}, which introduces the energy propagation mechanism in heterogeneous graphs, as well as the training and testing stages. Algorithm~\ref{alg:OODHG} outlines the overall process of OODHG.

\subsection{Learning Node Representations in Heterogeneous Graphs}
The HGNN used in this paper is SeHGNN~\cite{SEHGNN}. SeHGNN employs a single-layer long meta-path structure. Specifically, given a heterogeneous graph \( \mathcal{G} = \{ \mathcal{V}, \mathcal{E}, \phi, \psi \} \), the target type \( t \) and a set of given meta-paths \(\Phi\), the model generates multiple feature matrices with different semantics \( \textbf{X}^{P_n} \). Each matrix \( \textbf{X}^{P_n} \) represents the feature vectors aggregated from neighboring nodes according to $n$-th meta-path $P_n$.

Given the disparate feature dimensions inherent to the various node types, the feature vectors derived from different meta-paths may also exhibit different dimensions, thereby precluding direct fusion. To address this issue, a multi-layer perceptron (MLP) is employed to project the feature vectors from various meta-paths into a shared feature space, yielding semantic features, expressed as \({\textbf{H}^{\prime}}^{{P_n}}=\mathrm{MLP}_{{P_n}}(\textbf{X}^{{P_n}})\). After obtaining the semantic projection vectors with uniform dimensions, a Transformer-based semantic fusion module is utilized to integrate their semantic information, producing output vectors \({\textbf{H}^{P_n}}\). By concatenating the output vectors, we derive the final embedding vector for the node of type \( t \):
\begin{equation}
\textbf{Z}_{t}=\mathrm{Concat}([\textbf{H}^{{P}_{1}}\|\textbf{H}^{{P}_{2}}\|\ldots\|\textbf{H}^{{P}_{n}}]).
\label{eq1}
\end{equation}

\begin{algorithm}[!t]
\caption{Out-of-Distribution Detection in Heterogeneous Graphs (OODHG)}
\label{alg:OODHG}
\begin{algorithmic}[1]
\REQUIRE {Heterogeneous graph $\mathcal{G} = (\mathcal{V}, \mathcal{E}, \phi, \psi)$, target type \( t \), meta-paths set $\Phi$, threshold $\tau$, propagation steps $k$}
\ENSURE {OOD detection result OOD\_nodes and ID node classification result ID\_label}

\STATE Input $\mathcal{G}$ into HGNN to obtain the output logits $\textbf{H}$.

\STATE \textbf{\%Energy Score Calculation}
\FOR{each node $v_i \in V$ and \( \phi(v_i)=t \)}
    \STATE Calculate energy score: 
    \(
    E_i = - \log \sum_{c=1}^{K} e^{h_{i}[c]}.
    \)
\ENDFOR

\FOR{each meta-path $P_n \in \Phi$}
    \FOR{each iteration $k$}
         \STATE Energy propagation: 
         
         $\mathbf{E}^{(k,P_n)} = \gamma \mathbf{E}^{(k-1,P_n)} + (1 - \gamma) \hat{A}\mathbf{E}^{(k-1,P_n)}$
    \ENDFOR
\ENDFOR
\STATE Compute the final energy score for each node $v_i$: 

\(
{E_i} = \frac{1}{|\Phi|} \sum_{P_n \in \Phi} E_i^{(k,P_n)}
\)

\STATE \textbf{\%OOD Detection and ID Node Classification}
\FOR{each node $v_i \in V$ and \( \phi(v_i)=t \)}
    \IF{$-E_i \leq \tau$}
        \STATE Add node $v_i$ to OOD\_nodes.
    \ELSE
        \STATE Classify ID node $v_i$ using softmax probabilities and add classification result to ID\_label.

    \ENDIF
\ENDFOR

\RETURN OOD\_nodes, ID\_label
\end{algorithmic}
\end{algorithm}

\subsection{Energy-based OOD Detection}
Currently, there has been significant research focused on OOD detection. However, most OOD detection methods rely on softmax scores, specifically the maximum classification probability given by the softmax function. This implies that samples with classification probability distributions closer to a uniform distribution are more likely to be considered OOD. For instance, given a heterogeneous graph, we aim to perform $K$-class classification on the nodes. Through the SEHGNN, we obtain the embedding $z_i$ for each node $v_i$, which is then passed through an MLP to produce $K$-dimensional logits \( h_i = \mathrm{MLP}(z_i) \). Here, $h_i[c]$ corresponds to the unnormalized logit of node $v_i$ belonging to the $c$-th class. The logits $h_i$ can be transformed into a classification probability distribution using the softmax function:
\begin{equation}
p(y_i|\mathcal{G}) = \frac{e^{h_i[y_i]}}{\sum_{c=1}^K e^{h_i[c]}}.
\label{eq2}
\end{equation}
The node is categorized as an OOD node if the maximum classification probability is lower than a threshold $\tau$. Methods based on softmax scores are simple and somewhat effective for OOD detection. However, research shows that due to model overconfidence, OOD samples can still have high softmax scores~\cite{Energy}. In contrast, energy scores can better differentiate between ID and OOD samples. Notably, Energy-based OOD detection methods have recently gained significant attention for their effectiveness~\cite{hofmann2024energy}.

Energy-based models (EBM)~\cite{EBM} establish an energy function that maps each input into a scalar, known as the energy score. The energy score for an input node \( v_i \) can be expressed as:
\begin{equation}
\label{eq3}
E_i = - \log \sum_{c=1}^{K} e^{h_{i}[c]}.
\end{equation}
A higher energy \( E_i \) indicates a higher probability that \( v_i \) is an OOD node, whereas a lower energy suggests an ID node. An OOD detector \({G}(v_i) \) can then be used to determine whether a given node is OOD. The decision rule is as follows:

\begin{equation}
\label{eq4}
{G}(v_i)=\left\{\begin{array}{ll}
1, & \text { if } \quad -E_i \leq \tau, \\
0, & \text { if } \quad -E_i>\tau,
\end{array}\right.
\end{equation}

\noindent where $\tau$ is a predefined threshold, and the function returns 1 if \( v_i \) is an OOD sample and 0 otherwise. Negative energy scores align with traditional definitions, where lower values indicate a higher likelihood of an OOD sample. 

By comparing Equation~\ref{eq2} and Equation~\ref{eq3}, it can be observed that the energy score directly reflects the absolute magnitude of the logits rather than their relative proportions. When the model outputs low confidence scores across all classes but with large numerical differences between logits components, softmax-based methods tend to assign high confidence, leading to misclassification. In contrast, the energy-based method does not suffer from this issue, effectively mitigating the misclassification problem. Furthermore, this approach does not require any modifications to the existing neural network architecture. 

It can be observed that the energy score for each sample can be computed without altering the network structure. Furthermore, energy-based OOD detection methods can still utilize the cross-entropy loss function without modification. Minimizing the cross-entropy loss enables the model to effectively classify samples while reducing the energy scores of the training samples. The following is the derivation of the loss function:

\begin{equation}
\label{eq5}
    \mathcal{L} = -\frac{1}{N} \sum_{i=1}^N  \log(p(y_i|\mathcal{G})) = \frac{1}{N} \sum_{i=1}^N\left(-h_i[y_i]+\log\sum_{c=1}^Ke^{h_i[c]}\right),
\end{equation}

\noindent where $N$ denotes the number of target type nodes, and $y_i$ represents the label for node \( v_i \). From this formulation, it can be observed that as the cross-entropy loss is minimized, the energy of the training data also decreases.

\subsection{Energy Propagation in Heterogeneous Graphs}
In contrast to images, graphs capture not only individual feature information for each sample but, more importantly, intricate structural relationships. These relationships are critical in many tasks. Node representation learning aggregates the features of neighboring nodes, enabling more robust representations through the effective utilization of structural information. In the task of OOD detection, the connection patterns between nodes and the global graph structure similarly provide valuable information, enhancing detection accuracy. Existing research has demonstrated combining entropy propagation with structural information on homogeneous graphs achieves superior results~\cite{EMP}. 

However, in heterogeneous graphs, the variety of node types and the complexity of relationships complicate the information propagation process. Unlike homogeneous graphs, where nodes of the same type are usually directly connected, heterogeneous graphs do not always follow this pattern. This difference makes traditional propagation methods for homogeneous graphs unsuitable for heterogeneous ones. Consequently, identifying similar neighboring nodes requires leveraging metapaths to guide the propagation process. In a manner similar to message passing and neighbor aggregation in GNNs, we propose an energy propagation method specifically tailored for heterogeneous graphs. First, we calculate the energy scores for all nodes using Equation \ref{eq3}. Then, we propagate these scores along different metapaths. For the selection of metapaths, we first refer to the setting in ~\cite{SEHGNN} and filter out meta-paths whose hop count (e.g., the hop count of PAP is 2) is less than a predefined maximum metapath hop. Then, we further select meta-paths where both endpoints belong to the target node type as candidate paths for energy propagation. For a given meta-path \( P_n \), the vector of energy scores for nodes in a heterogeneous graph after \( k \) iterations of energy propagation can be expressed as: 
\begin{equation}
\label{eq6}
\mathbf{E}^{(k, {P_n})}=\gamma \mathbf{E}^{(k-1, {P_n})}+(1-\gamma) \hat{A} \mathbf{E}^{(k-1, {P_n})},
\end{equation}

\noindent where \( \gamma \in (0,1) \) is the weight coefficient to balance the energy contribution between a node itself and its neighbors, preventing excessive smoothing that may lead to information loss during propagation. \( \hat{A} \) is the row-normalized form of the adjacency matrix of target node type, which can be derived by matrix computation with the following equation:

\begin{equation}
\label{eq7}
\hat{A}=\hat{A}_{t_{1}, t_{2}} \hat{A}_{t_{3}, t_{4}} \ldots \hat{A}_{t_{l-1}, t_{l}},
\end{equation}

\noindent where \( \hat{A}_{t_i, t_{i+1}} \) refers to the row-normalized adjacency matrix \( A_{t_i, t_{i+1}} \) between node types \( t_i \) and \( t_{i+1} \). Notably, in this formula, both \( t_1 \) and \( t_{l} \) must correspond to the target node type. Ultimately, the final energy score for node \( v_i \) is calculated by averaging energy scores across all meta-paths:

\begin{equation}
\label{eq8}
{E_i} = \frac{1}{|\Phi|} \sum_{{P_n} \in \Phi }E_i^{(k, {P_n})}.
\end{equation}

\subsection{Optimization}
In graph-based tasks, the availability of labeled data for training is often very limited. Therefore, selecting an appropriate loss function is crucial for model training. This section introduces the loss functions we employed. To accomplish the tasks of OOD detection and ID node classification, we employed energy loss for OOD detection and classification loss for ID node classification to guide the model training. The overall objective function is defined as follows:

\begin{equation}
\label{eq9}
\mathcal{L} =\alpha\mathcal{L}_{\text{C}} + (1-\alpha)\mathcal{L}_{\text{E}}.
\end{equation}
where $\alpha$ is the weight coefficient that controls the trade-off between two loss functions.
\textbf{Classification loss}. This approach employs cross-entropy loss, which, when minimized, ensures accurate classification of ID nodes and lowers the energy values of labeled nodes~\cite{Energy}. The following is the definition of the cross-entropy loss:
\begin{equation}
\label{eq10}
    \mathcal{L}_{\text{C}} = -\frac{1}{N} \sum_{i=1}^{N} \log(p(y_i|\mathcal{G})).
\end{equation}

\textbf{Energy loss}. As OOD nodes are present only in the test set, only ID node labels are available during the training phase. While minimizing cross-entropy loss can reduce the energy scores of ID nodes, there may still be instances where the energy scores of ID nodes are close to those of OOD nodes, thereby diminishing the efficiency of OOD detection. We introduce an energy loss function designed to enhance the discriminative power of the energy distribution between ID and OOD nodes to address this issue. The formulation of the energy loss is as follows:
\begin{equation}
\label{eq11}
\mathcal{L}_{\text{E}} = \frac{1}{N} \sum_{i=1}^{N} \left[ \max(0, E_i - m_{\text{in}}) \right]^2,
\end{equation}

\noindent where $m_{\text{in}}$ is the margin-based hyperparameter, and $E_i$ represents the final energy score of node $v_i$ obtained after energy propagation. We employ a squared hinge loss term to constrain the energy, penalizing ID nodes with energy scores $E_i > m_{\text{in}}$, thereby encouraging the energy scores of ID nodes $E_i$ to be as low as possible to distinguish them from OOD nodes.

\section{Experiments}
\subsection{Experimental Setup}

\textbf{Evaluation Metrics}. The OOD detection tasks in heterogeneous graphs consist of two main tasks: (1) OOD detection, and (2) ID node classification based on the results of the first task. Different evaluation metrics are used to effectively evaluate the performance of the model in these two tasks, ensuring a comprehensive evaluation of the model's capabilities in both tasks.

The first task is framed as a binary classification problem, with the goal of distinguishing between OOD and ID nodes. The performance of the model on this task is evaluated using the following metrics:

\begin{itemize}
    \item \textbf{Area Under the Receiver Operating Characteristic Curve (AUROC)}: This metric measures the model's ability to detect between OOD nodes. A higher AUROC implies that the model performs better at distinguishing between ID and OOD nodes.
    \item \textbf{Area Under the Precision-Recall Curve (AUPR)}: AUPR is particularly useful for imbalanced datasets, focusing on the trade-off between precision and recall for the OOD class.
    \item \textbf{False Positive Rate at 95\% True Positive Rate (FPR@95)}: This metric quantifies the rate at which ID nodes are misclassified as OOD nodes when the true positive rate is fixed at 95\%, reflecting the model's ability to control false positives.
\end{itemize}

The second task is a \( K+1 \) classification problem, where \( K \) are ID classes and 1 is the OOD class. The performance on this task is evaluated using the following metrics:

\begin{itemize}
    \item \textbf{Micro-F1}: The Micro-F1 score aggregates performance across all classes, giving equal weight to each instance. It provides an overall measure of precision and recall at the global level.
    \item \textbf{Macro-F1}: The Macro-F1 score calculates the F1 score for each class independently before averaging the results across every class. This metric is particularly effective for imbalanced datasets, ensuring that each class is treated equally in the evaluation.
\end{itemize}

\begin{table}[width=\linewidth,cols=5,pos=t]
\caption{Statistics of datasets.}
\label{tab:dataset}
\begin{tabular*}{\tblwidth}{@{} CCCCCC@{} }
\toprule
\footnotesize\textbf{Dataset} & \footnotesize\textbf {\#Nodes}& \footnotesize\textbf{\#Node types}&\footnotesize\textbf{\#Edges} & \footnotesize\textbf{\#Edge types} & \footnotesize\textbf{\#Classes}\\
\midrule
 {\texttt{DBLP}} & 26,218 & 4 & 239,566 & 3 & 4\\
{\texttt{ACM}} & 10,942 & 4 & 547,872 & 4 & 3\\
{\texttt{IMDB}} &21,420 & 4 & 86,642 & 3 & 5\\
\bottomrule
\end{tabular*}
\end{table}

\textbf{Datasets}. We conducted experiments using three widely recognized real-world datasets: \texttt{DBLP}, \texttt{ACM}, and \texttt{IMDB}~\cite{SimpleHGNN}. These datasets exhibit distinct characteristics and are extensively utilized in heterogeneous graph-related tasks. We will provide a detailed description of each dataset in the follow.

\begin{itemize}
\item \texttt{DBLP} is a well-known citation network that is widely used in heterogeneous graph analysis due to its clear semantic structure that reflect meaningful academic relationships. The node types include author: A, paper: P, term: T, and venue: V, with the target node type being author. The edge types include AP, PA, TP, PT, VP, and PV. The meta-path set used for energy propagation is \{APA\}.

\item \texttt{ACM} is another citation network that is particularly suitable for studying heterogeneous graphs due to its relevance to academic knowledge discovery. The node types include P: paper, author: A, subject: S, and term: T, with the target node type being paper. The node types include PA, AP, PS, SP, PT, TP, PcP (paper cites paper), and PrP (paper is referenced by paper). The meta-path set used for energy propagation is \{PP, PAP, PCP, PAPP, PCPP, PPAP, PPCP, PAPAP, PAPCP, PCPAP, PCPCP\}.

\item \texttt{IMDB} is a movie-related network that is chosen for its distinct domain and the ability to model complex interactions in entertainment data. The node types include M: movie, director: D, actor: A, and keyword: K, with the target node type being movie. The edge types include MD, DM, MA, AM, MK, and KM. The meta-path set used for energy propagation is \{MAM, MDM, MKM, MAMD, MDMD, MAMAM, MAMDM, MAMKM, MDMAM, MDMDM, MDMKM, MKMAM, MKMDM, MKMKM\}. 
\end{itemize}

The statistics of these datasets are shown in Table \ref{tab:dataset}. As in traditional semi-supervised node classification (SSNC) tasks, we select 24\% of the nodes for the training, 6\% for validation, and the rest for the test set. To align with the OOD detection task setting, one class is designated as the OOD class, excluded from the training phase and only present in the test set. The node classification task on the \texttt{IMDB} dataset is a multi-label classification problem, which requires special handling. Specifically, we retain the first two labels for each node in \texttt{IMDB} and convert them into a four-class one-hot encoding. For example, if a node's original labels are [0,0], the one-hot encoded labels become [1,0,0,0]. It is noteworthy that in multi-label classification, a single sample may belong to multiple categories simultaneously, and there may exist certain correlations or dependencies among these labels. The simplification from multi-label to single-label classification can result in the loss of significant label information. However, this approach was chosen to facilitate a more straightforward comparison with existing methods that primarily focus on single-label classification tasks.

\textbf{Baseline Methods}. The baseline methods are categorized into two types: (1) methods designed for graph data (OpenWGL~\cite{OpemWGL}, OODGAT~\cite{OODGAT}, GNNSafe~\cite{GNNSafe}), which are primarily developed for homogeneous graphs, and (2) classical methods for OOD detection (MSP~\cite{MSP}, ODIN~\cite{ODIN}, Energy~\cite{Energy}), which are widely used in image and text domains. To adapt graph-based methods to heterogeneous graphs, we convert heterogeneous graphs into homogeneous ones using different metapaths, as described in~\cite{HAN}, and report the best performance. For classical OOD detection methods, we replace their backbone networks with SeHGNN.

\begin{table*}[!t]
\centering 
\normalsize
\caption{OOD detection results. The best result is highlighted in bold, while the runner-up is marked with \underline{underline}. And \(\uparrow\) (\(\downarrow\)) indicates that the larger (smaller) values are better.}
\label{tab2}
\setlength{\tabcolsep}{3pt}

\resizebox{\textwidth}{!}{
\begin{tabular}{cccccccccc}
\toprule                           
& \multicolumn{3}{c}{\texttt{DBLP}}  & \multicolumn{3}{c}{\texttt{ACM}} & \multicolumn{3}{c}{\texttt{IMDB}} \\
 & {AUROC↑}  & {AUPR↑}  & {FPR@95↓}  & {AUROC↑} & {AUPR↑} & {FPR@95↓} & {AUROC↑} & {AUPR↑} & {FPR@95↓} \\ 
 \midrule
 {OpenWGL} & {81.45±0.77} & {91.98±0.78} & {80.21±2.09} &  73.96±0.66 & 81.36±0.49 & 85.98±1.08 & 50.58±0.70 & 65.22±0.86 & 97.39±0.29\\
 {OODGAT} & {75.78±0.64} & {89.61±0.97} & {72.50±1.45} &  75.04±1.14 & 82.34±0.62 & 81.92±1.88 & 49.18±0.73 & 61.31±0.78 & 97.75±0.73\\
 {GNNSafe} & {87.51±0.61} & {95.20±0.32} & {50.18±1.20} &  81.04±0.43 & 86.21±0.46 & 72.01±0.88 & 55.68±0.79 & 67.80±0.26 & 94.68±1.15\\
 \hline
{MSP} & {94.53±0.38} & 98.05±0.17 & {31.42±3.22} &  78.26±0.84 & 86.22±0.28 & 69.49±2.30 & 58.73±0.76 & 95.67±0.40 & 92.89±0.91 \\
{ODIN}  & 94.26±0.68 & 97.98±0.38  & 29.32±1.52 & 82.61±1.26 & 88.01±0.98  & 58.94±1.14 & 59.09±0.28 & 95.41±0.15 & 92.90±0.50\\
Energy  & \underline{95.47±0.53} & \underline{97.90±0.27} & \underline{16.20±0.60} & \underline{89.22±0.25} & \underline{94.62±0.44}  & \underline{47.70±0.59} & \underline{59.55±0.66} & \underline{95.92±0.09} & \underline{92.71±0.76} \\
\bottomrule 
OODHG(HAN) & {96.33±0.32} & {98.27±0.18} & {12.73±1.25} &  {93.43±0.23} & {95.13±0.15} & {20.65±0.92} & {59.49±0.45} & {95.88±0.24} & {92.70±1.04}\\
OODHG(Simple-HGN) & {97.74±0.19} & {98.68±0.08} & {8.83±0.73} &  {95.32±0.35} & {97.52±0.27} & {15.42±1.01} & {60.34±0.44} & {96.03±0.31} & {90.58±1.08}\\
OODHG(SeHGNN) & \textbf{98.11±0.21} & \textbf{99.03±0.16} & \textbf{5.99±1.06} &  \textbf{95.58±0.45} & \textbf{97.59±0.18} & \textbf{14.92±0.77} & \textbf{62.02±0.21} & \textbf{96.23±0.11} & \textbf{89.16±0.91}\\ \hline
\end{tabular}
\vspace{-3mm}

}
\end{table*}

\begin{table*}[!t]
\centering 
\footnotesize
\caption{K+1 node classification results. The best result is highlighted in bold, while the runner-up is marked with \underline{underline}. And \(\uparrow\) indicates that the larger values are better.}
\label{tab3}
\resizebox{!}{!}{
\begin{tabular}{ccccccc}
\toprule                          
& \multicolumn{2}{c}{\texttt{DBLP}}  & \multicolumn{2}{c}{\texttt{ACM}} & \multicolumn{2}{c}{\texttt{IMDB}}\\
 & {Micro-F1↑} & {Macro-F1↑} & {Micro-F1↑} & {Macro-F1↑} & {Micro-F1↑} & {Macro-F1↑}\\ 
 \midrule
{OpenWGL} &  70.84±0.71 &  70.63±0.56 & 63.08±0.52 & 68.01±0.64 & 39.89±0.78 &56.87±.0.41\\
{OODGAT} &  68.09±0.37 &  69.34±0.54 & 64.88±0.90 & 64.73±0.34 & 38.09±1.05 &55.97±0.40\\
{GNNSafe} &  80.39±0.43 &  81.07±0.49 & 70.23±0.34 & 70.74±0.84 & 44.09±0.78 &59.61±1.21\\
\hline 
{MSP} &  87.80±0.70 &  87.31±0.52 & 79.61±0.49 & 80.17±0.43&\underline{50.83±0.39} & 63.18±0.18 \\
{ODIN}  & 86.83±0.80 & 86.29±0.97 & 79.36±0.84 & 78.90±0.83 &  50.60±0.36 & 63.11±0.11\\
Energy    &\underline{88.97±0.42} & \underline{88.94±0.43}&\underline{81.42±0.82} &\underline{82.40±0.44} & 50.52±0.72 & \underline{65.30±0.43}\\
\bottomrule
OODHG(HAN)  & {86.97±0.51} & {86.54±0.49} & {82.67±0.34} & {83.25±0.37} & {48.73±0.42} & {60.34±0.35}\\
OODHG(Simple-HGN)  & {90.08±0.34} & {89.45±0.28} & {85.67±0.24} & {85.97±0.23} & {50.16±0.25} & {62.78±0.21}\\
OODHG(SeHGNN)  & \textbf{91.37±0.46} & \textbf{90.71±0.46} & \textbf{86.13±0.43} & \textbf{86.58±0.39} & \textbf{51.94±0.28} & \textbf{66.97±0.17}\\ \hline
\end{tabular}
\vspace{-3mm}

}
\end{table*}

\textbf{Implementation Details}. For all comparison methods, we used the same dataset splits and random seeds to ensure consistency. The parameter settings for the heterogeneous graph neural network (SeHGNN) followed the specifications in the original paper. The learning rate was set to $1 \times 10^{-3}$, with a maximum of 50 iterations, and the optimizer used was Adam. To compute the F1 score, a threshold $\tau$ needs to be set. We employed a straightforward approach for threshold selection: for the energy score threshold, we explored values between 1 and 2 with a step size of 0.05, while for the softmax score threshold, values from 0.5 to 0.9 were tested in increments of 0.05. We reported the corresponding best performance for each case. In \texttt{DBLP}, the energy score threshold was set to 1.45, the softmax score threshold to 0.5, and the margin hyperparameter $m_{in}$ to -3. In \texttt{ACM}, the energy score threshold was set to 1.35, the softmax score threshold to 0.6, and the margin hyperparameter $m_{in}$ to -2. In \texttt{IMDB}, the energy score threshold was set to 1.95, the softmax score threshold to 0.45, and the margin hyperparameter $m_{in}$ to -3.

\subsection{Comparative Results}

We conducted a detailed comparison between OODHG and two types of baseline methods to demonstrate the superior performance of our method. Additionally, we investigated the impact of different HGNN backbones on our method, comparing the performance of using HAN~\cite{HAN} and Simple-HGN~\cite{SimpleHGNN} as backbones. The experimental results reported in Tables~\ref{tab2} and Table~\ref{tab3}.

\textbf{OOD Detection.} In Table~\ref{tab2}, we compare OODHG and the two classes of baseline methods on the OOD detection task across the \texttt{DBLP}, \texttt{ACM}, and \texttt{IMDB}, using commonly employed evaluation metrics: AUROC, AUPR, and FPR@95. The results clearly indicate that our proposed method, OODHG, exhibits a substantial performance advantage over the benchmark methods across all datasets. Specifically, the average AUROC of OODHG improves by 2.64\%, 6.36\%, and 2.47\% compared to the best baseline method on the three datasets, while the average AUPR increases by 1.13\%, 2.97\%, and 0.31\%. Additionally, our method significantly reduces the FPR@95, with decreases of 10.21\%, 32.78\%, and 3.55\%, respectively. The performance of OODHG on the \texttt{DBLP} dataset is particularly noteworthy, likely due to its smaller number of nodes but a higher number of edges, resulting in richer structural information, which OODHG can effectively leverage. In summary, the results indicate the robustness and effectiveness of OODHG in addressing the OOD detection task.

\textbf{K+1 Node Classification.} Table~\ref{tab3} compares OODHG and the two classes of baseline methods on the K+1 node classification task, using standard evaluation metrics: Micro-F1 and Macro-F1. In this task, our method demonstrates consistent superiority over the benchmark methods across all datasets. Specifically, the average Micro-F1 of OODHG improves by 2.4\%, 4.71\%, and 1.42\% compared to the best baseline method on the three datasets, while the average Macro-F1 increases by 1.77\%, 4.18\%, and 1.67\%. Notably, the Macro-F1 is significantly higher than the Micro-F1 on the \texttt{IMDB} dataset, likely due to our handling of the dataset labels, resulting in class imbalance. Furthermore, the straightforward method for selecting $\tau$ may prevent the selection of an optimal $\tau$ value, potentially resulting in performance below the optimal level.

Overall, OODHG is a specialized OOD detection method tailored for heterogeneous graphs, which effectively integrates both the feature information of all nodes and the intricate topological structure within heterogeneous graphs. As a result, it significantly outperforms all baseline methods in both OOD detection and K+1 node classification tasks in heterogeneous graph scenarios. Methods designed for homogeneous graphs (e.g., openWGL, OODGAT, and GNNSafe) exhibit suboptimal performance when applied to heterogeneous graphs. The primary reason is that these methods struggle to adapt to heterogeneous graph data. They often fail to effectively preserve the feature information of non-target-type nodes and the structural information of the graph. As a result, model performance declines significantly. Although classical OOD detection methods (e.g., MSP, ODIN, and Energy) attempt to accommodate heterogeneous graph data by introducing HGNN as the backbone network, their failure to fully leverage the unique structural properties of heterogeneous graphs results in inferior performance compared to OODHG. These experimental results not only validate the effectiveness of our model but also underscore its robustness and superiority in handling complex graph-structured data. Additionally, it is worth noting that among all the baseline methods, the Energy-based approach performs notably well in the OOD detection task, suggesting that energy-based models are better suited for addressing the challenging OOD detection task compared to softmax probability-based methods. Moreover, the results indicate that changing the HGNN model can impact detection performance. Stronger HGNNs outperform other models in both OOD detection and node classification tasks, which may be attributed to the fact that the energy score calculation and classification rely on the quality of node representations. This finding suggests that when an HGNN is able to learn more expressive node representations, it can considerably enhance the performance of OOD detection. 

    
    

\begin{figure*}[!t] 
    \centering
    \begin{subfigure}[b]{0.32\textwidth} 
        \centering
        \includegraphics[width=\textwidth]{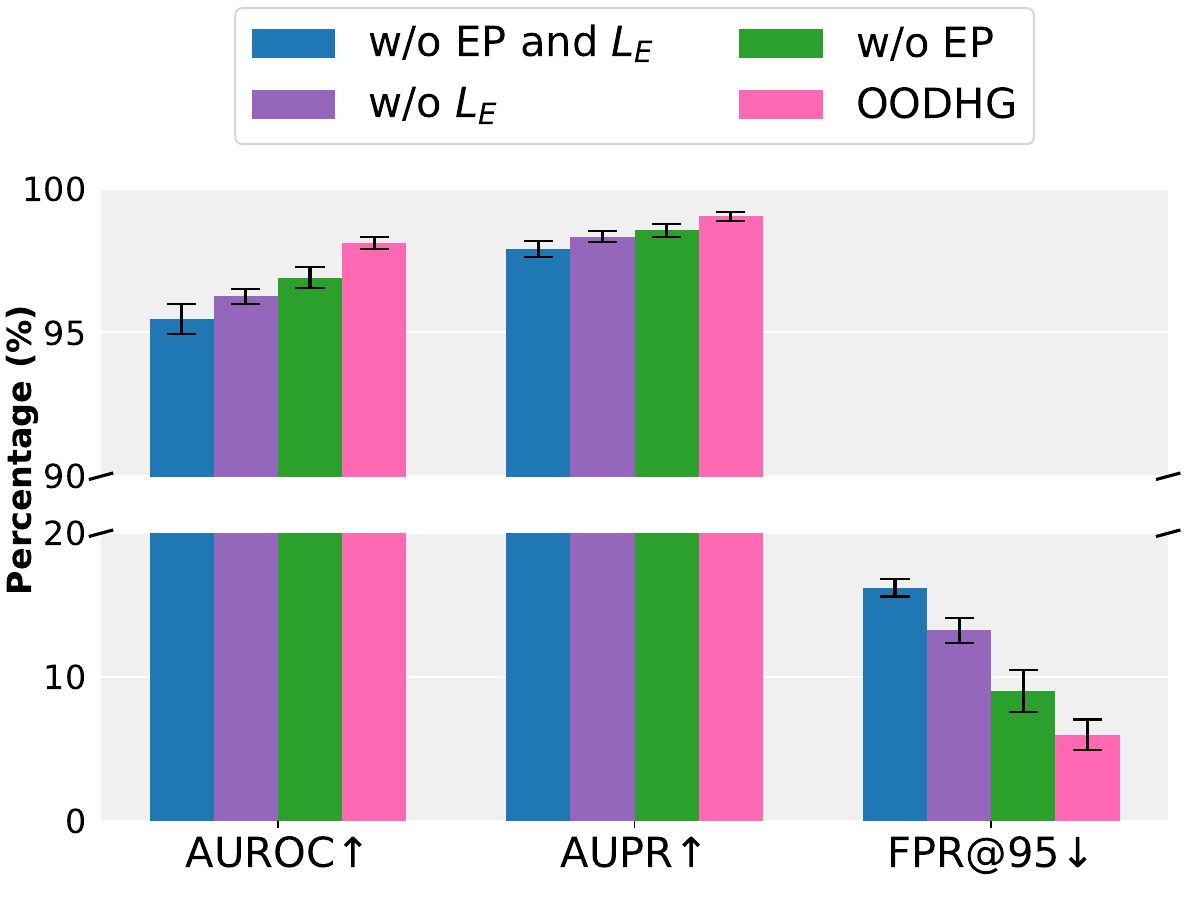}
        \caption{Ablation study for \texttt{DBLP}.}
        \label{fig:ablation_DBLP}
    \end{subfigure}
    \hspace{0mm} 
    \begin{subfigure}[b]{0.32\textwidth} 
        \centering
        \includegraphics[width=\textwidth]{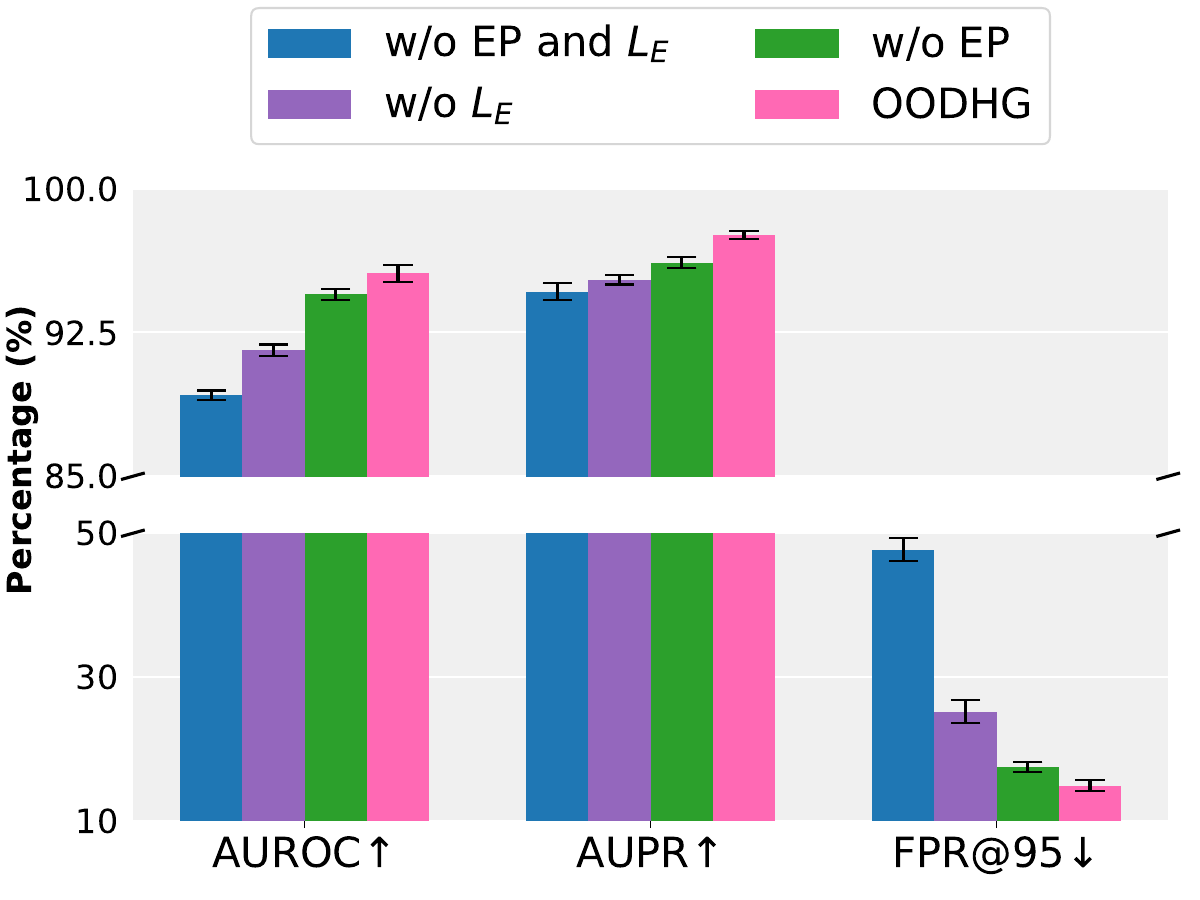}
        \caption{Ablation study for \texttt{ACM}.}
        \label{fig:ablation_acm}
    \end{subfigure}
    \hspace{0.0mm} 
    \begin{subfigure}[b]{0.32\textwidth} 
        \centering
        \includegraphics[width=\textwidth]{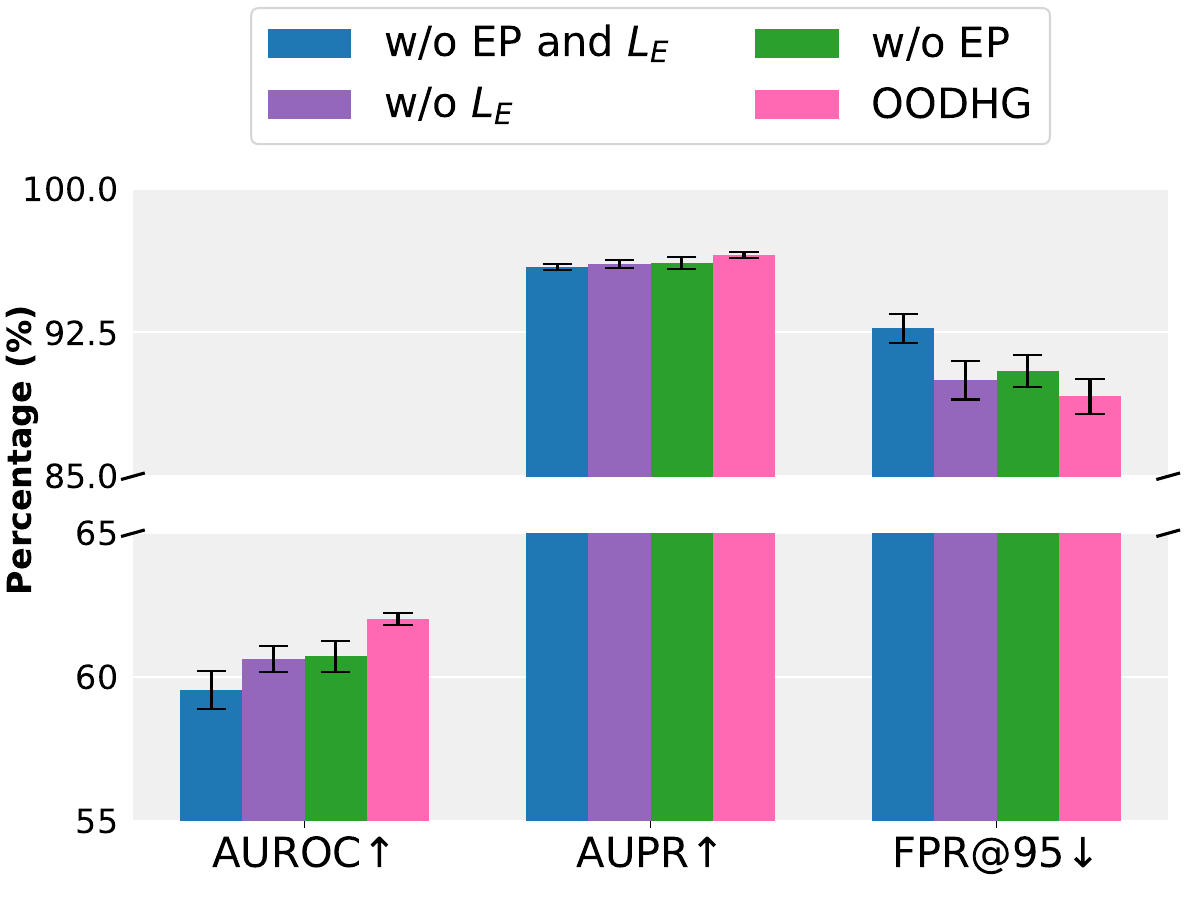}
        \caption{Ablation study for \texttt{IMDB}.}
        \label{fig:ablation_imdb}
    \end{subfigure}
    
    \caption{Ablation study on three datasets.}
    \label{fig:3}
\end{figure*}

\subsection{Ablation Study}
In this section, we present the results of ablation experiments conducted to validate the effectiveness of the energy propagation (EP) mechanism and energy loss ($L_E$) in heterogeneous graphs. The model is evaluated under four different configurations as follows: (1) w/o EP and $L_E$, where neither energy propagation nor energy loss is applied, and only the classification loss is considered; (2) w/o $L_E$, where energy propagation is applied, but the model is trained solely with classification loss; (3) w/o EP, where energy loss is incorporated, but energy propagation is not applied; (4) OODHG, where both energy propagation and energy loss are employed to guide the training process. The results are presented in Figure~\ref{fig:3}, with Figures~\ref{fig:3}(a), (b), and (c) corresponding to \texttt{DBLP}, \texttt{ACM}, and \texttt{IMDB}, respectively.

By comparing the results of (1) and (2), it is evident that the energy propagation mechanism significantly enhances AUROC and AUPR by leveraging the abundant structural information inherent in the heterogeneous graph, while also reducing FPR@95. A comparison between (1) and (3) shows that the incorporation of the energy loss function leads to improved OOD detection performance, as it assigns higher energy scores to ID nodes. Furthermore, the improvements are more pronounced on the \texttt{DBLP} dataset, potentially due to its smaller dataset size, which enables the energy loss function to guide the training process more effectively. A comparison among (2), (3), and (4) indicates that when both energy propagation and the energy loss function are employed simultaneously to guide the model's training, the distinction between ID and OOD nodes becomes more pronounced, leading to superior performance in the OOD detection task compared to using either energy propagation or the energy loss function alone.

Our ablation study has demonstrated that EP and $L_E$ improve OOD detection performance. However, the impact of these components varies across different datasets. The effectiveness of EP depends on information propagation between nodes. In DBLP (small-scale, densely connected), EP can fully exploit rich structural information, enhancing OOD detection performance. In contrast, in IMDB (large-scale, sparsely connected), the limited information flow weakens the effectiveness of EP. LE enhances the distinction between ID and OOD nodes by reducing the energy scores of ID nodes, but its effectiveness depends on the quality of node representations. In DBLP and ACM, where HGNN learns more discriminative node representations, LE significantly improves performance. However, in IMDB, the label simplification may degrade the quality of learned node representations, thereby limiting the contribution of LE to performance improvement.

    

\subsection{Visualization Analysis}
\begin{figure*}[!t]    
    \centering
    \begin{subfigure}[b]{0.32\textwidth}
        \centering
        \includegraphics[width=\textwidth]{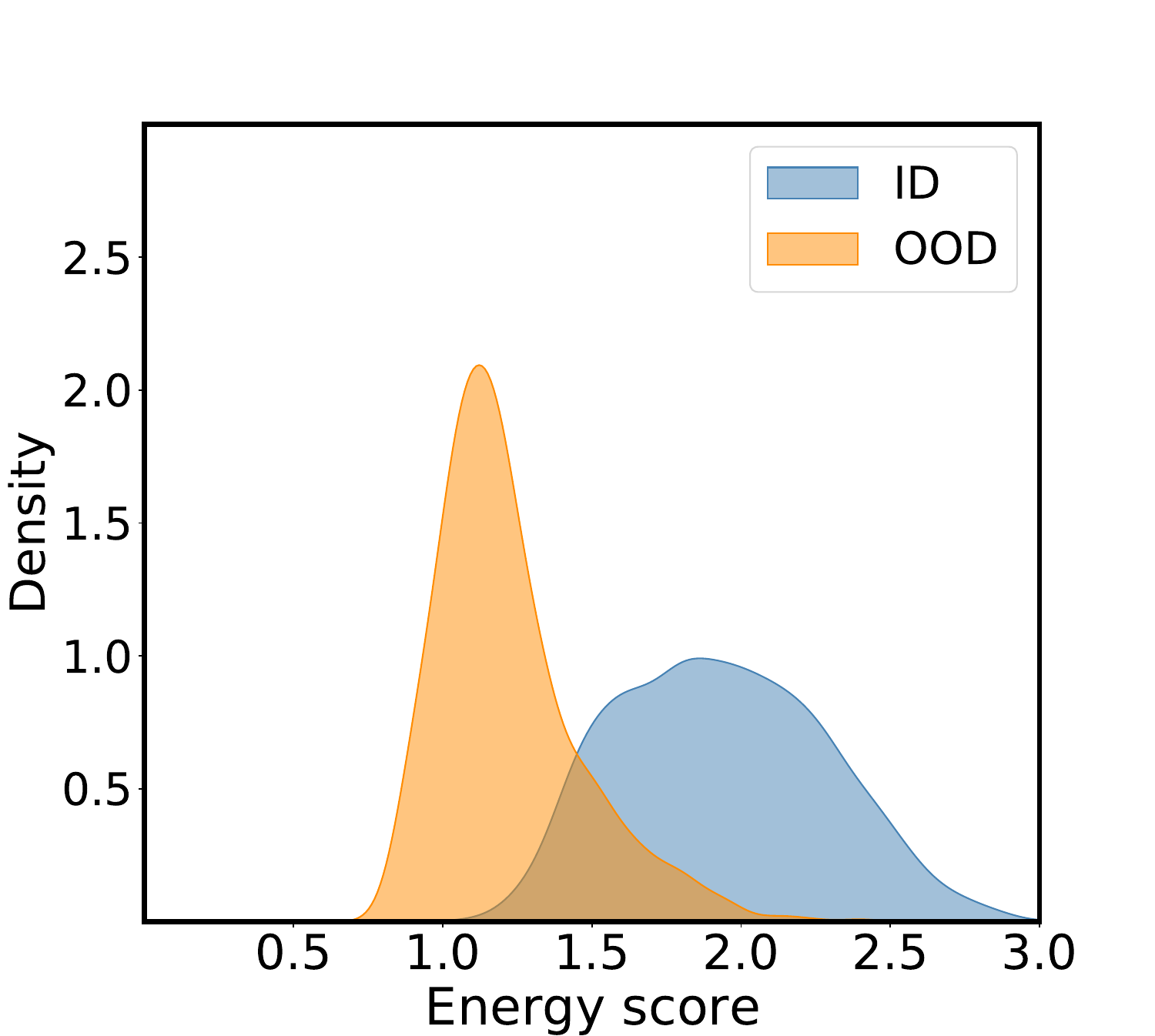}
        \caption{}
        \label{fig:subfig1}
    \end{subfigure}
    \vspace{-0mm}
    \hspace{-0.0cm}
    \begin{subfigure}[b]{0.32\textwidth}
        \centering
        \includegraphics[width=\textwidth]{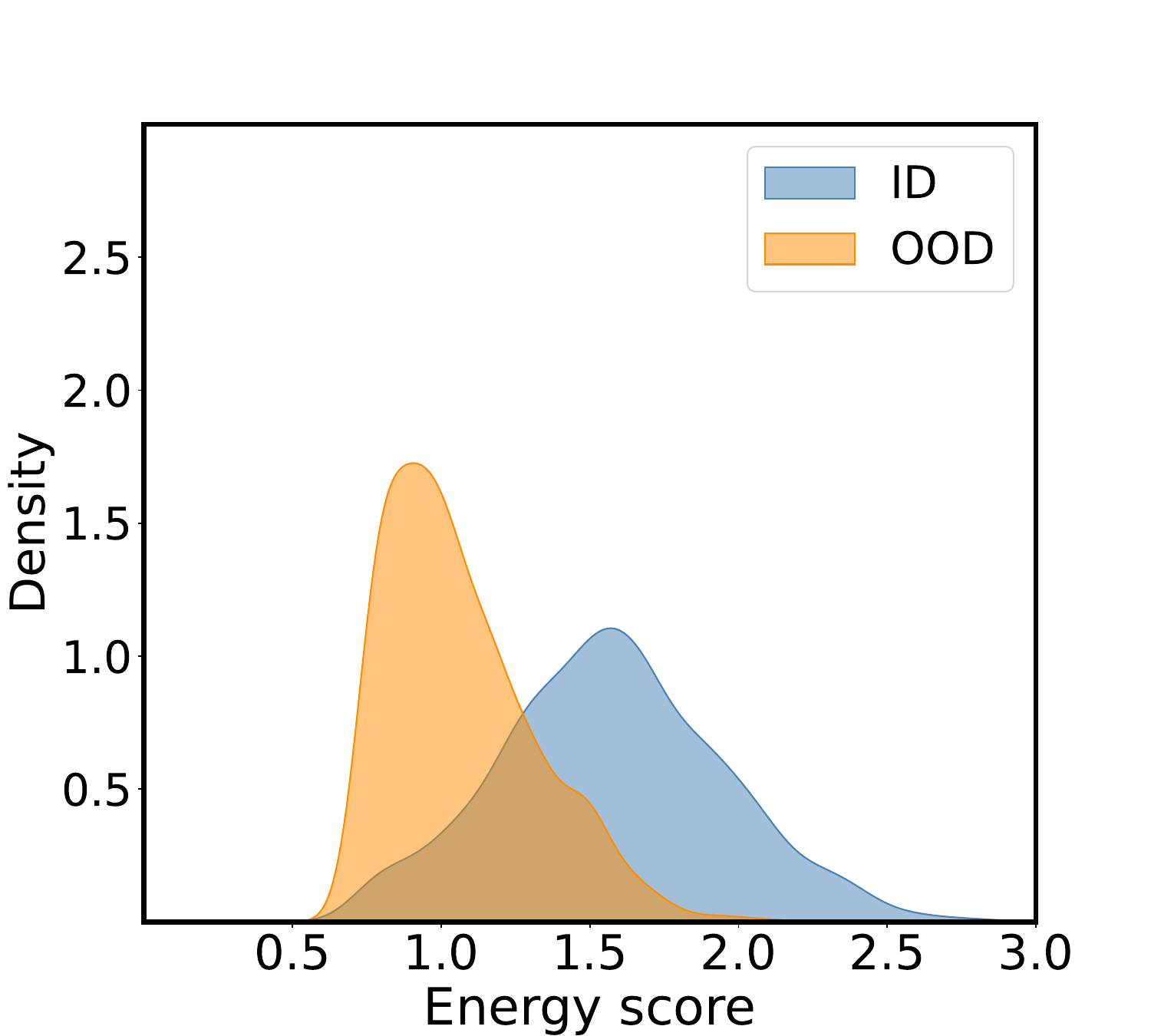}
        \caption{}
        \label{fig:subfig2}
    \end{subfigure}
    \vspace{-0mm}
    \begin{subfigure}[b]{0.32\textwidth}
        \centering
        \includegraphics[width=\textwidth]{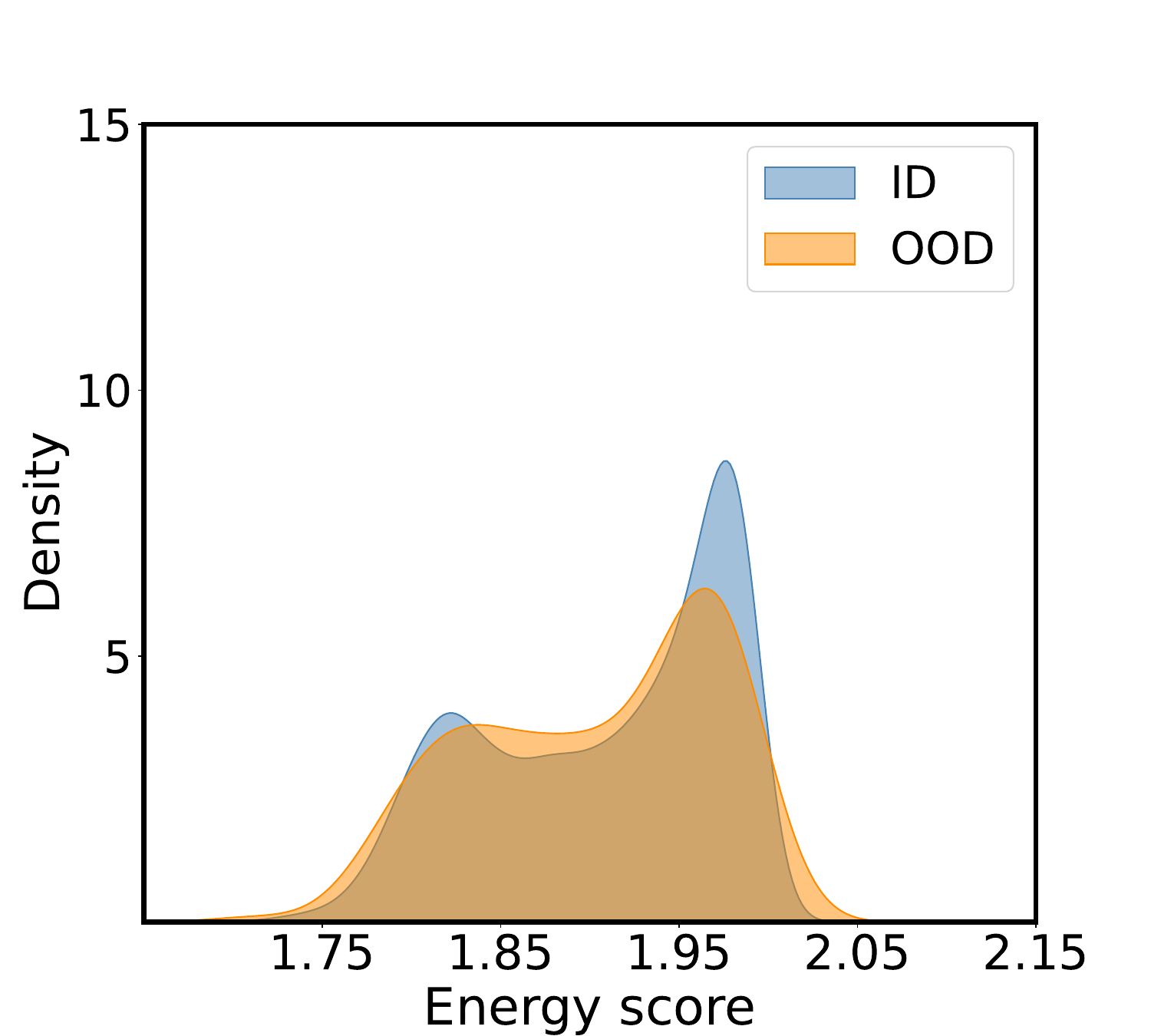}
        \caption{}
        \label{fig:subfig3}
    \end{subfigure}
    \hspace{-0.0cm}
    \begin{subfigure}[b]{0.32\textwidth}
        \centering
        \includegraphics[width=\textwidth]{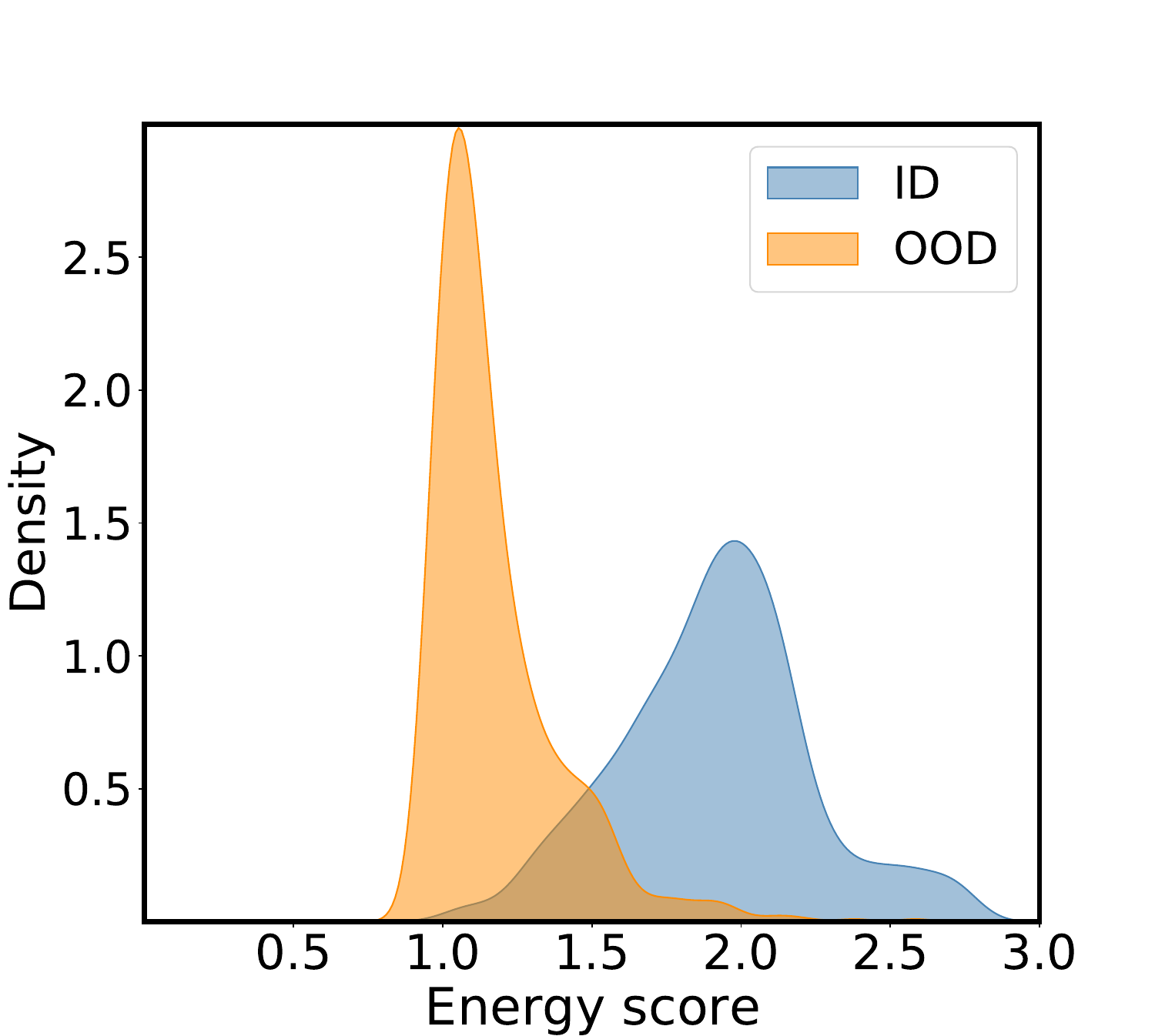}
        \caption{}
        \label{fig:subfig4}
    \end{subfigure}
    \begin{subfigure}[b]{0.32\textwidth}
        \centering
        \includegraphics[width=\textwidth]{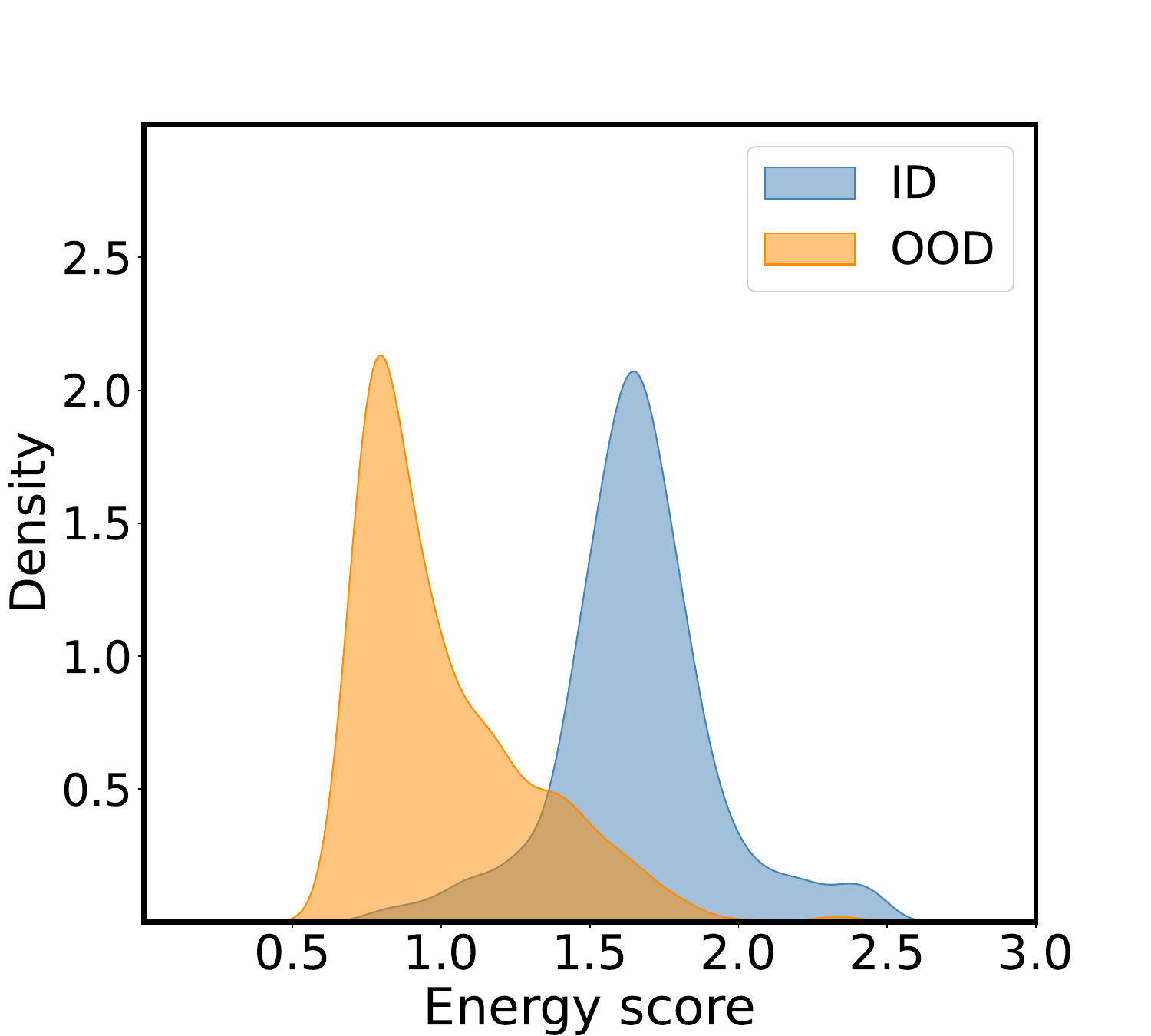}
        \caption{}
        \label{fig:subfig5}
    \end{subfigure}
    \hspace{-0.0cm}
    \begin{subfigure}[b]{0.32\textwidth}
        \centering
        \includegraphics[width=\textwidth]{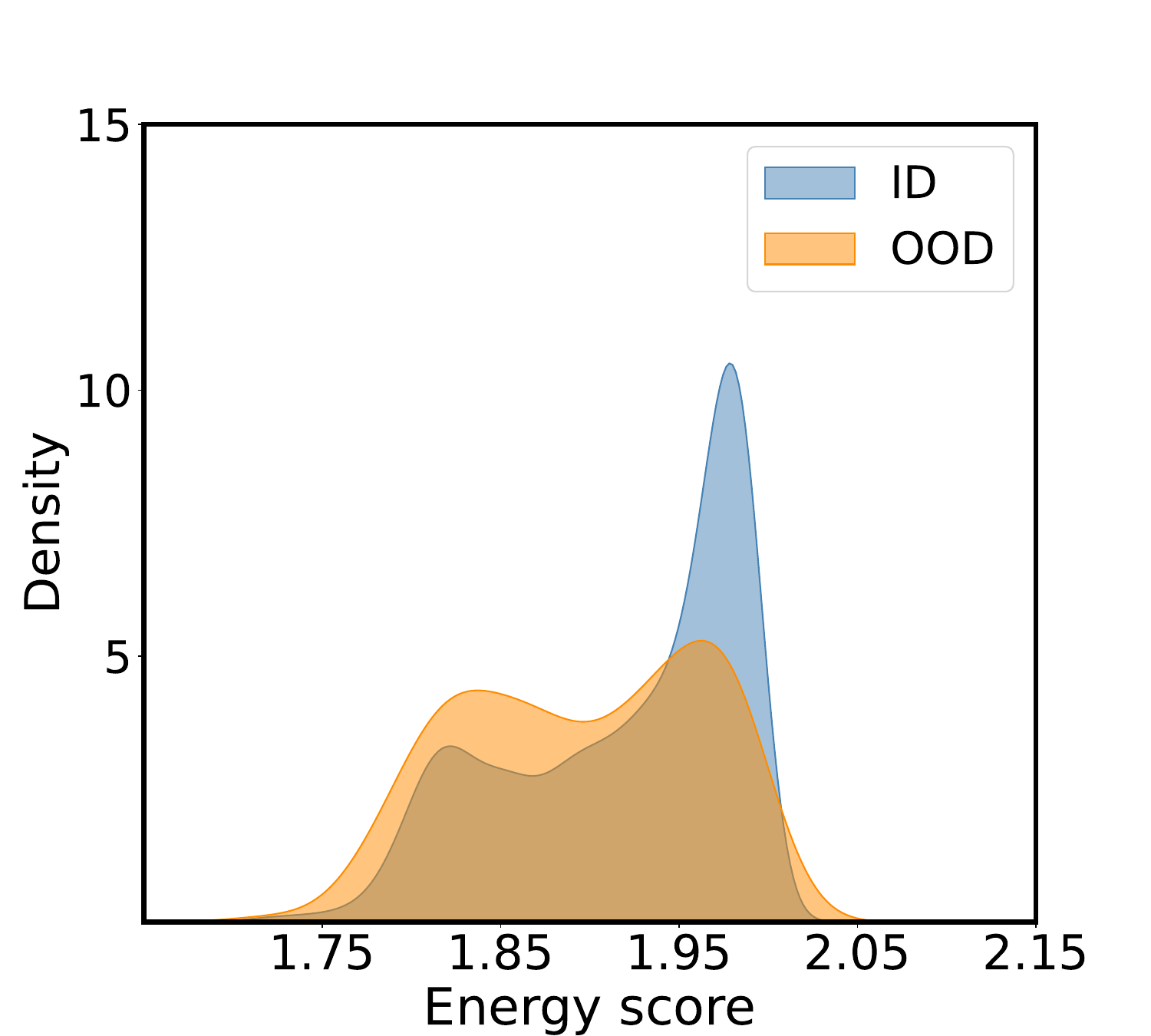}
        \caption{}
        \label{fig:subfig6}
    \end{subfigure}
    \caption{The visualized results of the energy scores on three datasets.}
    \label{fig:4}
    \vspace{-3mm}
\end{figure*}

To provide a more intuitive demonstration of the energy propagation mechanism on heterogeneous graphs, we conducted additional visual analysis of the energy distribution for both OOD and ID nodes across three datasets. In the \texttt{DBLP} dataset, Figures~\ref{fig:4}(a) and (d) show the energy distributions of nodes before and after the application of energy propagation, respectively. Similarly, for the \texttt{ACM} dataset, Figures~\ref{fig:4}(b) and (e) present the energy distributions before and after energy propagation. For the \texttt{IMDB} dataset, Figures~\ref{fig:4}(c) and (f) illustrate the energy distributions before and after energy propagation.

From the observations, we see that in the \texttt{DBLP} and \texttt{ACM} datasets, the energy distributions of ID and OOD nodes exhibit significant overlap before the application of energy propagation. However, after energy propagation, this overlap is greatly reduced, and the peak density of the ID and OOD distributions increases noticeably, indicating a clearer distinction between the two. In contrast, for the \texttt{IMDB} dataset, ID and OOD nodes show similar energy distributions, with limited improvement after energy propagation. This limited enhancement may be attributed to the simplification of the multi-label classification task into a single-label classification task in the \texttt{IMDB} dataset, which results in the loss of some label information and negatively impacts the model's performance.

\begin{figure*}[!t]    
    \centering
    \begin{subfigure}[b]{0.32\textwidth}
        \centering
        \includegraphics[width=\textwidth]{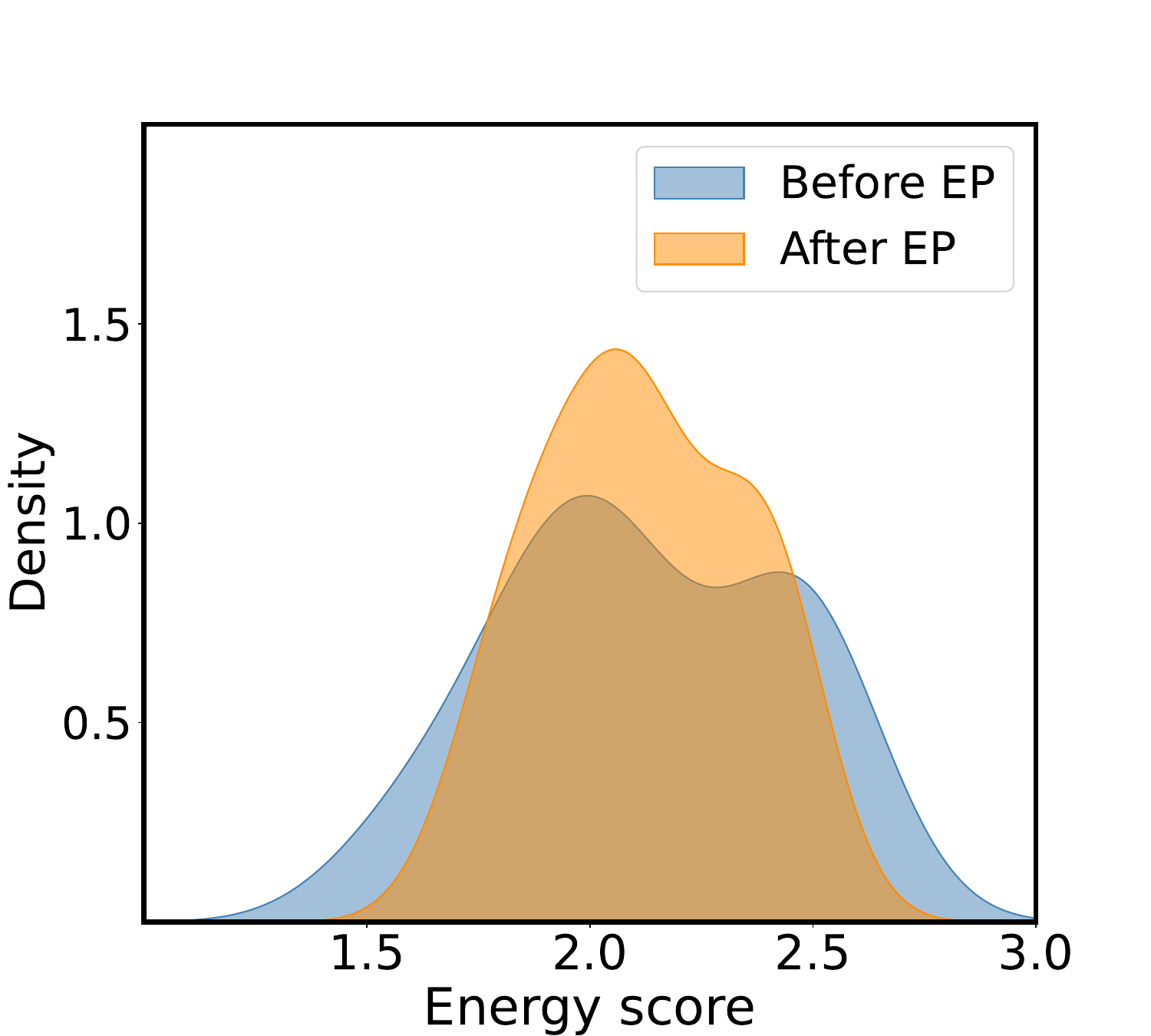}
        \caption{\texttt{DBLP}}
        \label{fig5:subfig1}
    \end{subfigure}
    \vspace{-0cm}
    \hspace{-0.0cm}
    \begin{subfigure}[b]{0.32\textwidth}
        \centering
        \includegraphics[width=\textwidth]{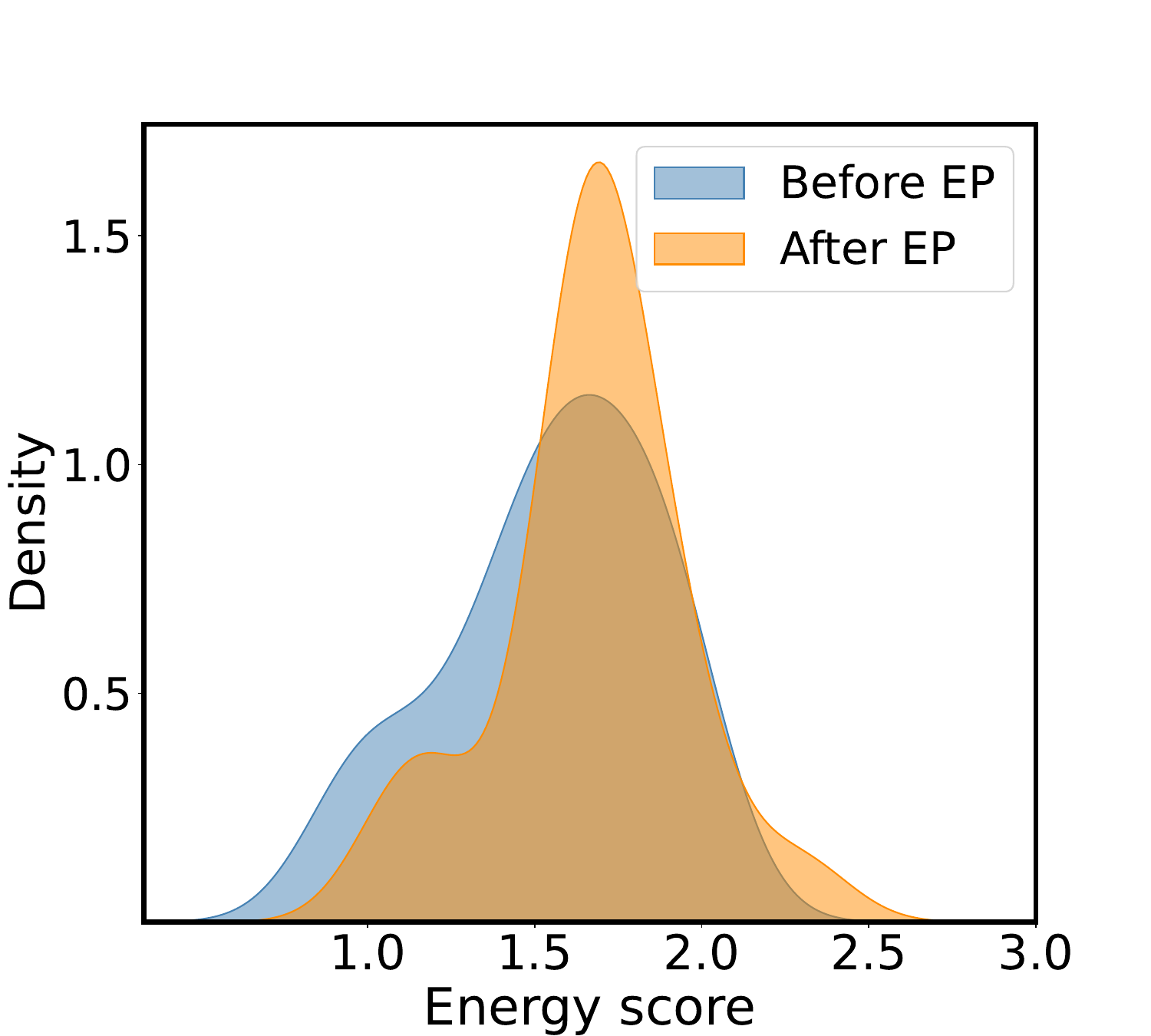}
        \caption{\texttt{ACM}}
        \label{fig5:subfig2}
    \end{subfigure}
    \vspace{-0cm}
    \begin{subfigure}[b]{0.32\textwidth}
        \centering
        \includegraphics[width=\textwidth]{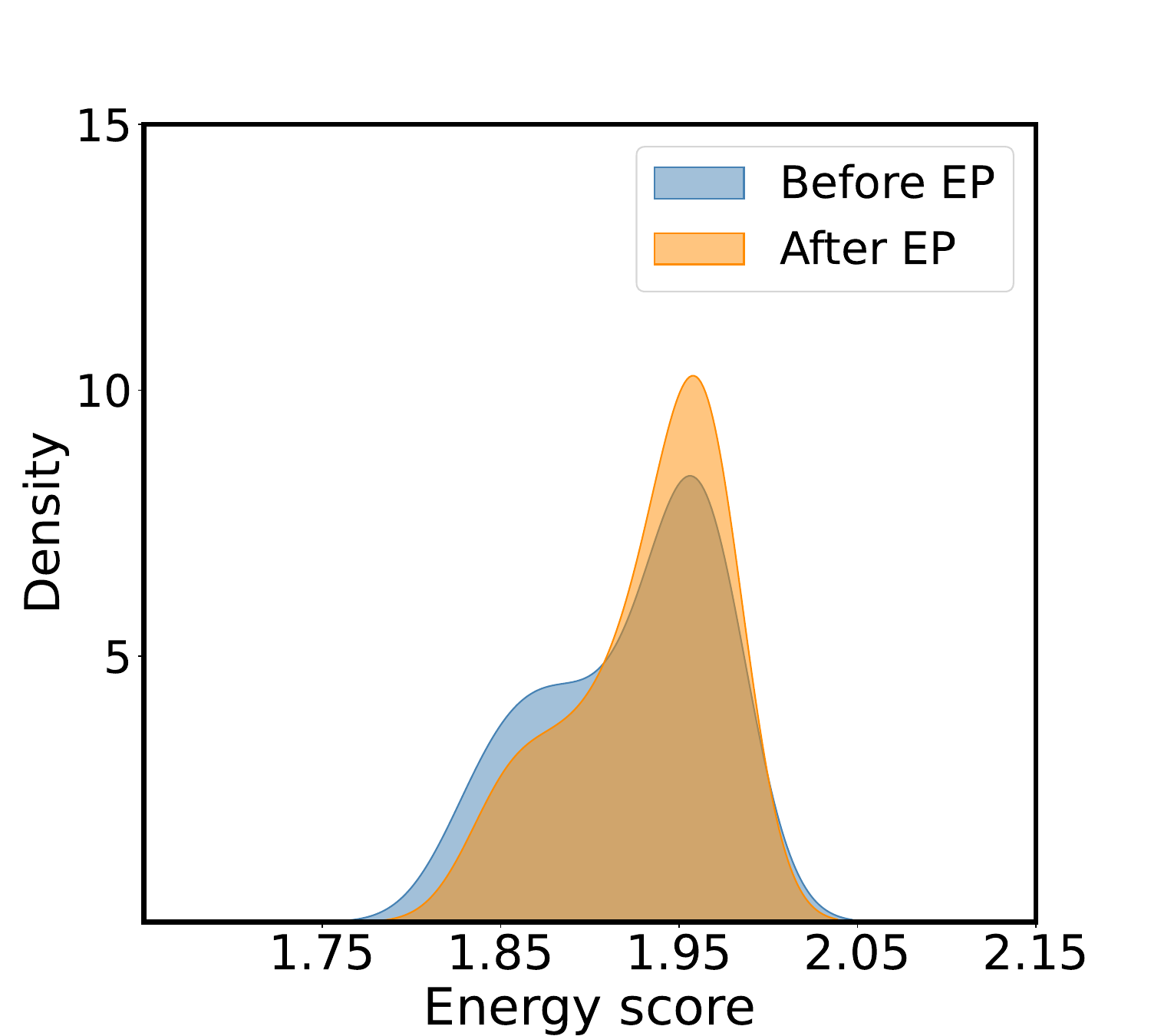}
        \caption{\texttt{IMDB}}
        \label{fig5:subfig3}
    \end{subfigure}
    \caption{Visualization of energy distribution of one class of nodes on three datasets.}
    \label{fig:5}
    \vspace{-3mm}
\end{figure*}

To intuitively demonstrate the effectiveness of the energy propagation mechanism, we visualize the energy distribution of a specific class of ID nodes in each dataset before and after energy propagation. As shown in Figure \ref{fig:5}, the experimental results indicate that, after applying energy propagation, the energy distribution of these nodes exhibits a significant clustering effect, and their average energy values are effectively increased. This optimization in energy distribution enhances the distinction between ID and OOD nodes in terms of energy characteristics, thereby validating the theoretical expectations of the energy propagation mechanism.

\subsection{Parameter Analysis}
We performed a parameter sensitivity analysis on the \texttt{DBLP} dataset to evaluate the impact of the energy propagation weight coefficient $\gamma$, the number of propagation steps $K$, the weight coefficient $\alpha$ of the loss function, and the margin hyperparameter $min_m$ on model performance.

\begin{figure*}[!tbp]
    \centering
        \begin{subfigure}[b]{0.42\textwidth}
            \centering
            \includegraphics[width=\textwidth]{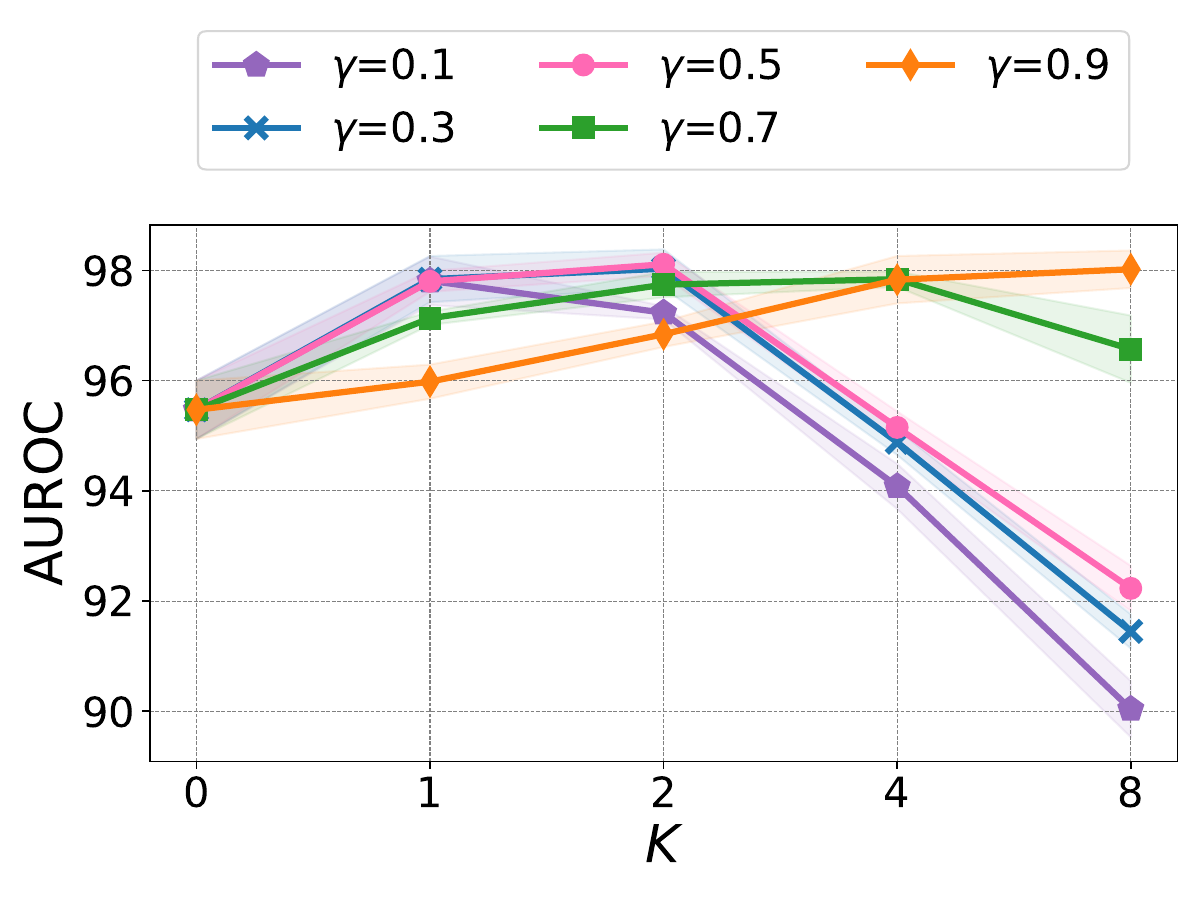}
            \caption{}
            \label{para:subfig1}
        \end{subfigure}
        \hspace{-0.0cm}
        \begin{subfigure}[b]{0.42\textwidth}
            \centering
            \includegraphics[width=\textwidth]{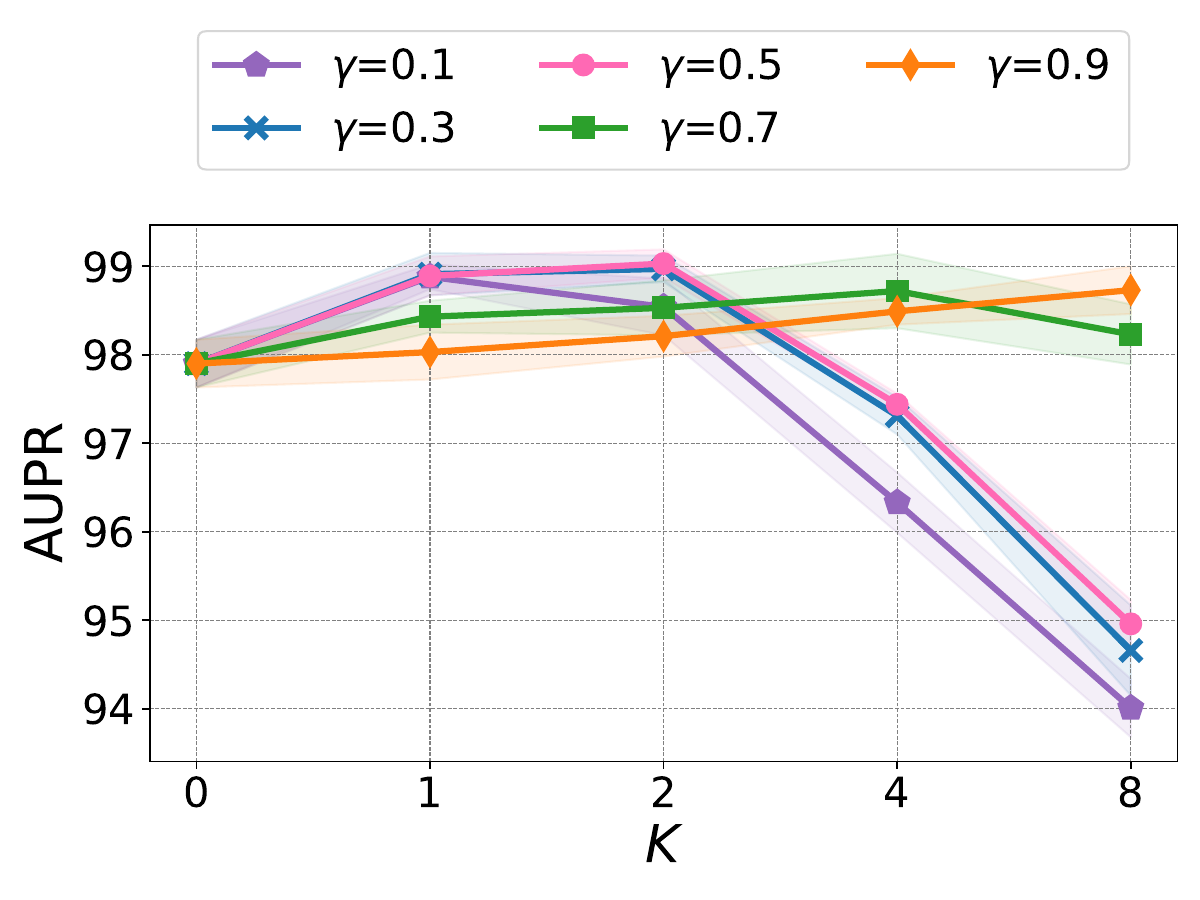}
            \caption{}
            \label{para:subfig2}
        \end{subfigure}
        \vfill
        \begin{subfigure}[b]{0.42\textwidth}
            \centering
            \includegraphics[width=\textwidth]{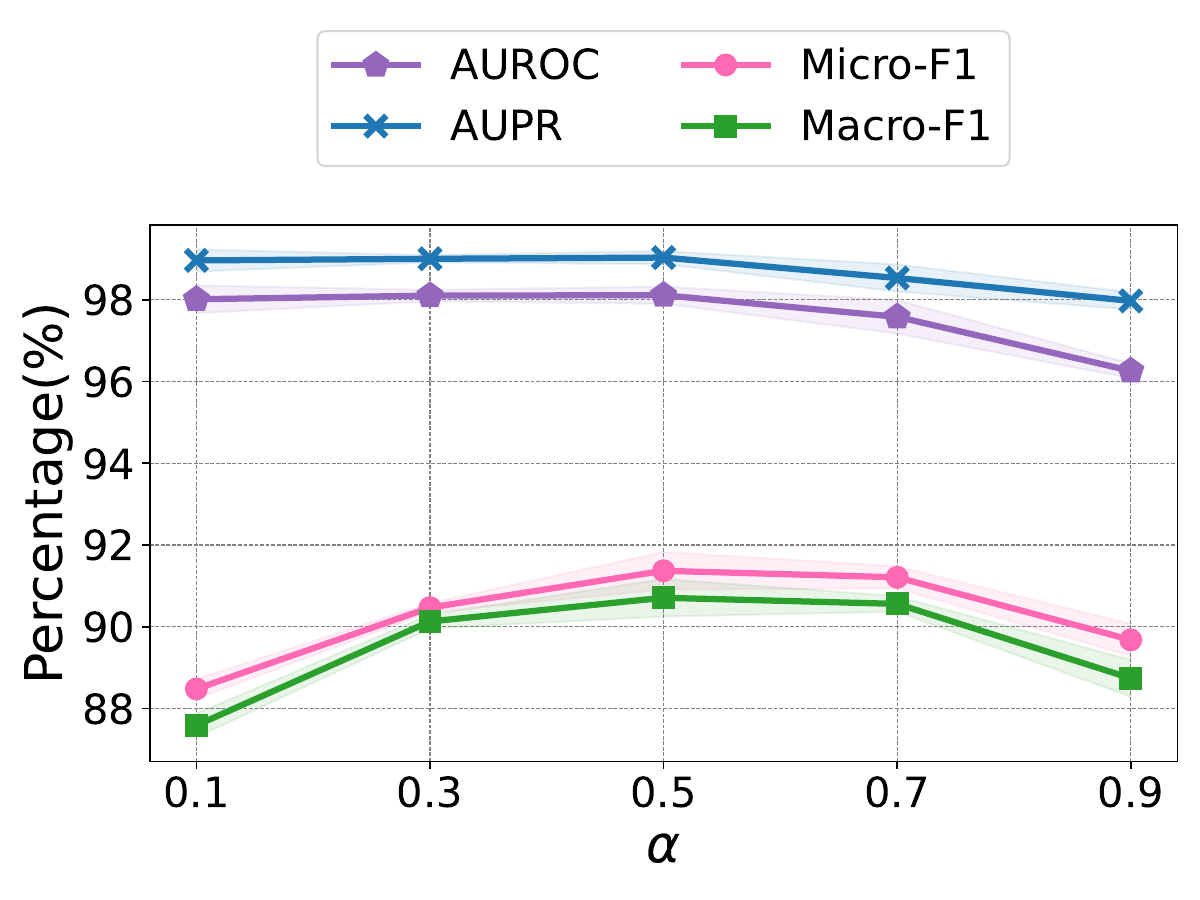}
            \caption{}
            \label{para:subfig3}
        \end{subfigure}
        \hspace{-0.0cm}
        \begin{subfigure}[b]{0.42\textwidth}
            \centering
            \includegraphics[width=\textwidth]{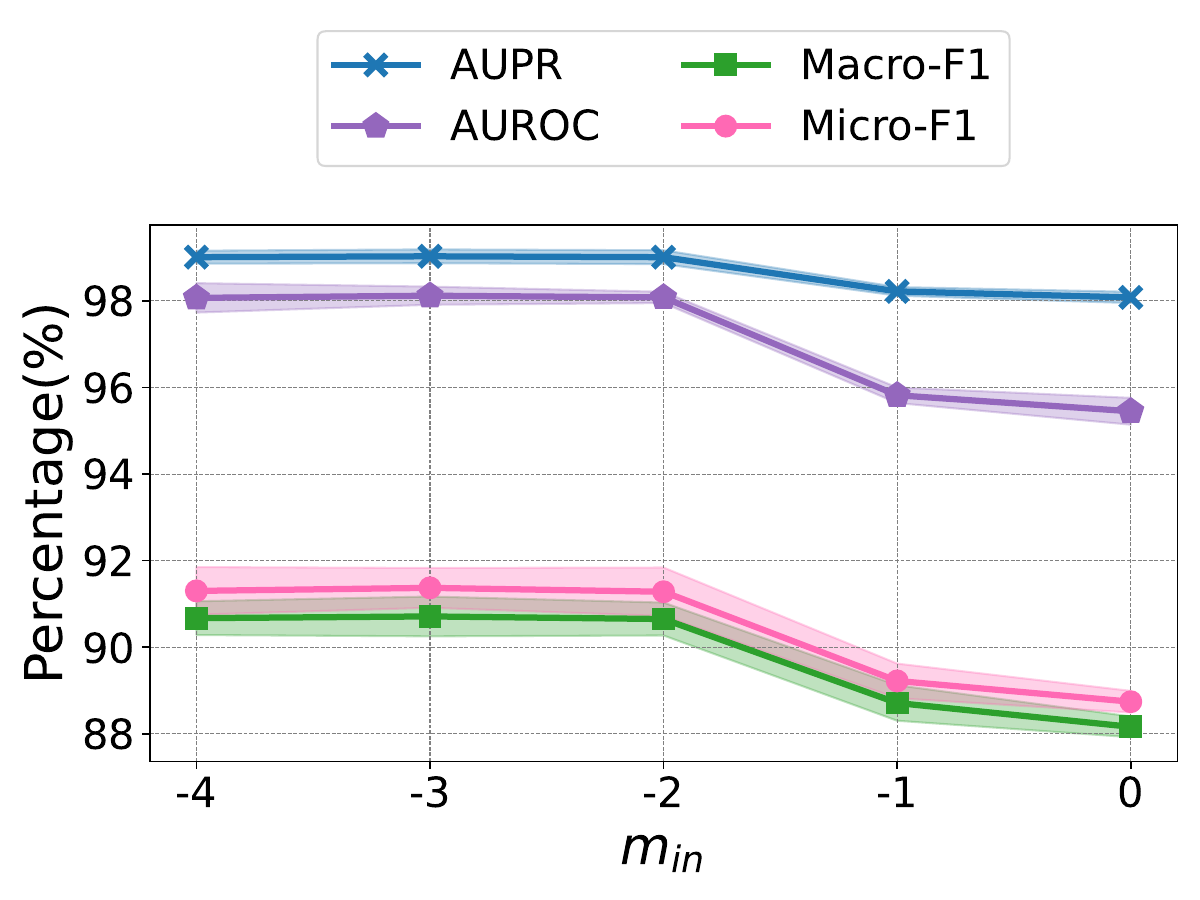}
            \caption{}
            \label{para:subfig4}
        \end{subfigure}
    \caption{Hyperparameter analysis on \texttt{DBLP}. (a)(b)Analysis of propagation weight coefficient $\gamma$ and propagation steps K. (c)Analysis of the loss function weight coefficient $\alpha$. (d)Analysis of Marginal Hyperparameters $m_{in}$. }
    \label{fig:6}
    \vspace{-3mm}
\end{figure*}

\textbf{Energy Propagation Weight Coefficient and Steps.} Figures~\ref{para:subfig1}~\ref{para:subfig2}   present the results, which evaluate the effects of $\gamma$ and $K$ on model performance. AUROC and AUPR were used as evaluation metrics.

For the analysis of the impact of $\gamma$: When $\gamma$ is too large, the weight of energy scores assigned to neighboring nodes during energy propagation decreases, and the model almost exclusively considers the energy score of the node itself, resulting in limited performance improvement. Conversely, when $\gamma$ is too small, the energy score assigned to the node itself becomes insufficient, leading to a performance drop. Therefore, selecting a moderate value of $\gamma$ is critical for achieving optimal model performance.

For the analysis of the impact of the propagation steps $K$: From the figure, it can be observed that as $K$ increases, model performance gradually improves in the initial stages. This suggests that a sufficient amount of energy propagation allows the model to fully leverage the structural information of the graph, thereby enhancing the OOD detection performance. However, when $K$ becomes too large, energy propagation may lead to overly smooth energy distributions among nodes, weakening the model’s ability to capture local structural information and thus impairing performance. Therefore, the choice of $K$ requires a balance between propagation depth and information dilution, with an optimal number of steps being essential for achieving the best performance.

\textbf{Loss Function Weight Coefficient.} Figures~\ref{para:subfig3} presents the results evaluating the impact of $\alpha$ on model performance using AUROC, AUPR, Micro-F1, and Macro-F1. As $\alpha$ decreases, the influence of energy loss $L_e$ increases, enhancing the energy score difference between ID and OOD nodes and improving OOD detection. However, an excessively small $\alpha$ weakens the cross-entropy loss $L_c$, potentially degrading ID classification performance. The optimal $\alpha$ depends on dataset characteristics. For datasets where OOD samples are highly distinct, a smaller $\alpha$ enhances detection. Conversely, when OOD samples closely resemble ID samples, a moderate $\alpha$ balances OOD detection and ID classification. Empirical tuning, such as grid search or validation set optimization, is essential for achieving the best performance across different datasets.

\textbf{Margin Hyperparameter.} Figures~\ref{para:subfig4} present the results evaluating the impact of $m_{in}$ on model performance using AUROC, AUPR, Micro-F1, and Macro-F1. Optimal performance is achieved when $m_{in} = -3$, and for $m_{in} \leq -2$, performance remains stable at near-optimal levels. While a smaller $m_{in}$ theoretically lowers energy scores for ID nodes, improving OOD detection, excessive reduction (e.g., $m_{in} = -4$) leads to overfitting, degrading generalization. The optimal $m_{in}$ depends on dataset characteristics. For datasets where ID and OOD distributions are highly separable, a more negative $m_{in}$ can enhance OOD detection without significantly impacting ID classification. However, for datasets where ID and OOD samples are closely distributed, an overly small $m_{in}$ may push ID node energy scores too low, leading to instability. Therefore, $m_{in}$ should be carefully tuned based on dataset properties, with empirical validation or grid search helping to determine the best range.

\subsection{Time Complexity Analysis}
This section evaluates the time complexity of OODHG. Assume that the graph contains \( r \) types of nodes, with the number of target node types being \( N_{\text{target}} \), and the number of nodes of each type denoted as \( N_1, N_2, \dots, N_r \). \( |\Phi| \) represent the number of meta-paths, \( l \) represent the length of each meta-path, and the algorithm be executed for \( k \) iterations.
In each iteration, the energy propagation process begins with the computation of the target node's adjacency matrix \( \hat{A} \). For a given meta-path \( P_n \), the computation of \( \hat{A} \) involves the multiplication of several matrices. The time complexity for each matrix multiplication is \( O(N_{t_1} \cdot N_{t_2} \cdot \dots \cdot N_{t_l}) \), where \( N_{t_1}, N_{t_2}, \dots, N_{t_l} \) are the sizes of the node types involved in the meta-path \( P_n \). The energy matrix propagation can be performed via the matrix multiplication \( \hat{A} \mathbf{E}^{(k-1)} \), where \( \mathbf{E}^{(k-1)} \) is the energy matrix from the previous iteration. The time complexity of this propagation process is \( O(|\Phi| \cdot N_{\text{target}}^2) \), where \( |\Phi| \) is the number of meta-paths and \( N_{\text{target}} \) is the number of target node types.

In practice, it is not necessary to recompute the adjacency matrix for the target nodes in each iteration; instead, the adjacency matrix \( \hat{A} \) for each meta-path \( P_n \) can be precomputed and stored. Therefore, after \( k \) iterations, the total time complexity  is:

\begin{equation}
    O( k \cdot |\Phi| \cdot N_{\text{target}}^2 )
\end{equation}

\begin{figure}[!tbp]
  \centering
  \includegraphics[width=\linewidth]{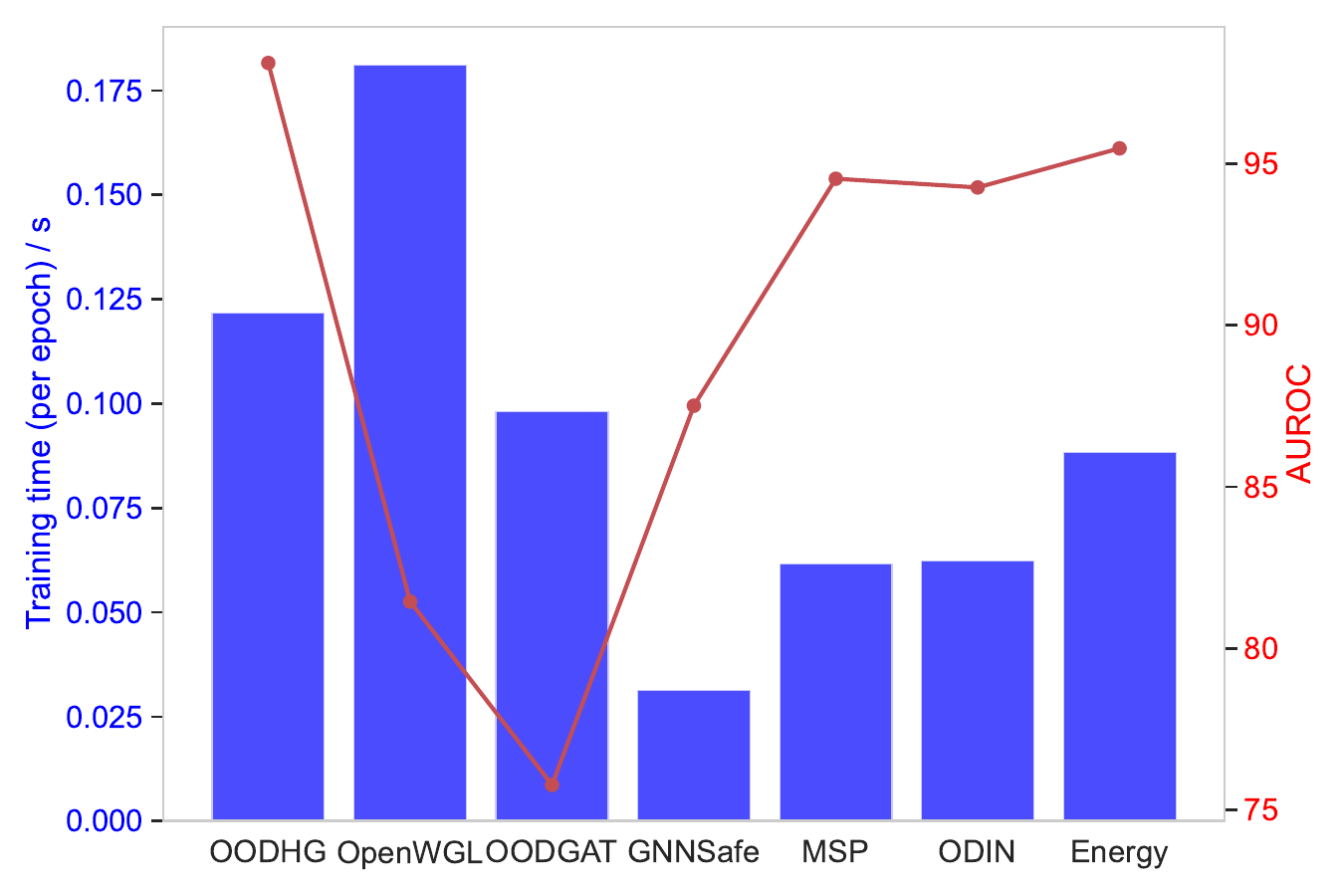}
  \caption{AUROC and time consumption of OODHG and baseline methods on the DBLP dataset.}
  \label{fig:time}
  \vspace{-3mm}
\end{figure}

To validate the feasibility of the OODHG model in real-world applications, this study compares the training time of OODHG with baseline methods on \texttt{DBLP}. As shown in Figure~\ref{fig:time}, the experimental results indicate that OODHG achieves a significant improvement in OOD detection performance without notably increasing the training time.

\section{Conclusion}
This paper proposes a novel approach, OODHG, for addressing the challenging task of OOD detection in heterogeneous graphs. Heterogeneous graphs contain complex structural information due to diverse node and edge types. Traditional methods designed for homogeneous graphs fail to capture this complexity, making them unsuitable for OOD detection. To overcome this limitation, we propose a meta-path-based energy propagation mechanism that fully leverages the rich structural information inherent in heterogeneous graphs. Furthermore, we incorporate an energy constraint to enhance the distinction between ID and OOD nodes. Our extensive experimental evaluations demonstrate that OODHG consistently outperforms state-of-the-art baseline methods, underscoring the critical role of meta-path-guided propagation in enhancing OOD detection performance. Despite the encouraging experimental results achieved in this study, several limitations remain, which provide some directions for future research. First, the fixed propagation parameters may not achieve optimal performance across different datasets. Future research could explore dynamic parameter adjustment mechanisms to further enhance the robustness of OOD detection. Second, the current method primarily focuses on static heterogeneous graphs, whereas real-world graph data often exhibit dynamic evolutionary characteristics. Therefore, extending the OODHG framework to dynamic heterogeneous graphs could be a valuable direction for future work. These improvements will contribute to enhancing the generalizability and practical applicability of the proposed approach.

\printcredits

\bibliographystyle{cas-model2-names}

\bibliography{cas-refs}





\end{document}